%% file: main_arxiv_version.tex
\documentclass{scrarticle}
\usepackage[utf8]{inputenc}
\usepackage[a4paper, left=2.5cm, top=2.5cm, right=2.5cm, bottom=2.5cm]{geometry}
\usepackage{xcolor}
\usepackage{tikz}
\usepackage{natbib}
\usepackage{amssymb}
\usepackage{amsmath}
\usepackage{tabularx, ragged2e}
\usepackage{subcaption}
\usepackage{kbordermatrix}
\usepackage{hyperref}
\usepackage[colorinlistoftodos, textsize=tiny]{todonotes}
\usepackage[croatian, english]{babel}
\usepackage{array}
\newcolumntype{Z}{>{\raggedright\let\newline\\\arraybackslash\hspace{0pt}}X}
\usepackage[htt]{hyphenat}
\usepackage{multirow}

\title{How trial-to-trial learning shapes mappings in the mental lexicon: Modelling Lexical Decision with Linear Discriminative Learning}
\author{Maria Heitmeier\footnote{maria.heitmeier@uni-tuebingen.de}, Yu-Ying Chuang and R. Harald Baayen\\Quantitative Linguistics, University of Tübingen}
\date{}

\newcommand{\chat}{$\hat{\mathbf{c}}$}

\begin{document}

\maketitle

\begin{abstract}
    Trial-to-trial effects have been found in a number of studies, indicating that processing a stimulus influences responses in subsequent trials. A special case are priming effects which have been modelled successfully with error-driven learning \citep{marsolek2008antipriming}, implying that participants are continuously learning during experiments. This study investigates whether trial-to-trial learning can be detected in an unprimed lexical decision experiment. We used the Discriminative Lexicon Model \citep[DLM;][]{baayen2019discriminative}, a model of the mental lexicon with meaning representations from distributional semantics, which models error-driven incremental learning with the Widrow-Hoff rule. We used data from the British Lexicon Project \citep[BLP;][]{keuleers2012british} and simulated the lexical decision experiment with the DLM on a trial-by-trial basis for each subject individually. Then, reaction times were predicted with Generalised Additive Models (GAMs), using measures derived from the DLM simulations as predictors. We extracted measures from two simulations per subject (one with learning updates between trials and one without), and used them as input to two GAMs. Learning-based models showed better model fit than the non-learning ones for the majority of subjects. Our measures also provide insights into lexical processing and individual differences. This demonstrates the potential of the DLM to model behavioural data and leads to the conclusion that trial-to-trial learning can indeed be detected in unprimed lexical decision. Our results support the possibility that our lexical knowledge is subject to continuous changes.
\end{abstract}

\newpage

\section{Introduction}
When going through our daily lives, we are constantly confronted with new information. What we see, hear and feel continuously updates our internal model of the world. This continuous learning shapes how we perceive, process, learn and react to the world \citep[e.g.][]{bennett2015single,diedrichsen2010use,nassar2010approximately,oreilly2018deep, oreilly2021deep,    ramscar2014myth,ramscar2016learning, ramscar2017mismeasurement}. Learning does not only change our perception at a general level, but it also has immediate consequences for how we react to the world given what we have just perceived or experienced. Experimentally, this effect can be observed for example in repetition priming: after processing some information, when similar information is encountered again, it is processed more easily, which usually results in e.g. higher accuracy compared to non-repeated information \citep[see][for a review]{mcnamara2005semantic, roediger1993implicit}. Analogously, it has been found across many domains that the opposite is also true: if a repeated or similar stimulus is followed by a different outcome, processing is impaired \citep[an effect often referred to as ``antipriming''; overview in][]{marsolek2008antipriming}.

It has been found in recent work that priming effects can be modelled with a simple error-driven learning rule, called the Rescorla-Wagner learning rule \citep{baayen2020modeling,hoppe2022exploration,marsolek2008antipriming,nixon2021prediction,oppenheim2010dark,Rescorla1972}. Error-driven learning, as modelled by the Rescorla-Wagner rule, assumes that when perceiving an input (often referred to as a \textit{cue}), activations of outcomes are predicted \citep[the terminology of cues and outcomes follows][]{Danks2003}. Then, the error between the actual outcome and its predicted activation are computed, and the mapping from the cue to the observed outcome is strengthened accordingly. Mappings from the cue to all other outcomes that were activated but not observed are weakened. This mechanism accounts for repetition priming: successful processing of cue $a$ and outcome $A$ results in the strengthening of the connection between cue $a$ and outcome $A$. As a consequence, when the same cue $a$ is encountered again, the outcome $A$ is activated more strongly, thus reducing error rates and processing time. At the same time, error-driven learning also provides an account of antipriming: Connection strengths to other outcomes which are not present in the learning event (e.g. to $B$) are weakened. As a consequence, if cue $a$ is presented again, outcome $B$ will be activated less and processing $B$ is impaired.

The Rescorla-Wagner rule has recently been applied successfully to language learning and subsequently in many areas of psycholinguistics. In its simplest forms, it has been used to account for issues in (second) language acquisition \citep{arnon2012granularity, ellis2006language, ellis2006selective,  ellis2010bounds,milin2017discrimination,milin2017learning} and ageing research \citep{ramscar2014myth, ramscar2017mismeasurement}, as well as semantic priming \citep{oppenheim2010dark}, morphological processing \citep{baayen2011amorphous,baayen2016frequency,milin2017discrimination}, learning of symbolic knowledge \citep{ramscar2010effects}, genre-decision making \citep{milin2020approaching}, the U-shaped learning of irregular English plurals in children \citep{ramscar2007linguistic, ramscar2013error} and early infant sound acquisition \citep{nixon2021prediction}.

Modelling priming with the Rescorla-Wagner rule assumes that the learning taking place during the processing of the prime changes the way in which the target is subsequently processed. If learning takes place in priming paradigms, from prime to target, then it likely also occurs in other tasks. Indeed, a number of previous studies have identified inter-trial effects in various paradigms both outside of \citep[e.g.][]{allenmark2021inter,gilden2001cognitive, jones2006recency, jones2013sequential,  palmeri2015experimental} and within psycholinguistics, specifically the lexical decision task. In a lexical decision task, participants have to decide whether a presented stimulus is an existing word in their language or not. Lexical decision is traditionally employed to probe representation and processing in the mental lexicon. One line of research found that global composition of stimuli in lexical decision experiments systematically changed reaction times \citep[e.g.][]{dorfman1988list, ferrand1996list, wagenmakers2008diffusion}. A different line of research focused on the effects of immediately preceding trials (often termed ``first-order sequential effects''). For example, it was found that the lexicality (i.e. word/nonword) of trial $n-1$ can influence the reaction time in trial $n$ \citep{lima1997sequential}. Characteristics of the stimuli other than lexicality can also have an influence: \citet{balota2018dynamic} found a four-way interaction between degradation and lexicality of the previous and current stimulus on reaction times, and \citet{perea2003sequential} found that if trial $n$ is a nonword or a low-frequency word, its reaction time is influenced by the frequency of the word in the previous trial, whereas if the stimulus in trial $n$ is a high-frequency word, there is no such influence. Various computational models have been developed to capture such inter-trial effects \citep[e.g.][]{allenmark2021inter, jones2006recency, jones2013sequential}. For example, the mathematical account for modelling inter-trial effects in two-answer-forced-choice tasks by \citet{jones2013sequential} uses previous stimuli categories and their repetition pattern to predict reaction times on subsequent trials. At an even more stimulus-specific level, early research demonstrated repetition priming in lexical decision: if a stimulus was shown repeatedly, reaction times became shorter \citep{forbach1974repetition, scarborough1977frequency}.

A number of computational studies have modelled trial-to-trial learning. Theories such as ACT-R \citep{anderson1998actr} can model the learning and forgetting of stimuli during experiments. It has also been shown that trial-to-trial learning can be modelled with the Rescorla-Wagner or related learning rules. \citet{oppenheim2010dark} studied semantic priming effects in a naming task, using an incremental learning model. Other studies showed that the learning of serial patterns \citep{tomaschek2022keys} and event-related potentials (ERPs) when listening to sequences of tones \citep{lentz2021temporal} can be predicted with a Rescorla-Wagner learning model. However, these models view representations in such learning tasks as mostly categorical. For example, both ACT-R and the model by \citet{oppenheim2010dark} treat words' forms as units, disregarding any effects that orthographical similarity might have on inter-trial learning. This disregards that similarity at a subcategorical level is the essence of the anti-priming effect of \citet{marsolek2008antipriming}, and underlies many of the results reported for example by \citet{ramscar2007linguistic, ramscar2013error}. Moreover, ACT-R models forgetting as a function of time \citep{vanrijn2003modeling}, without explicitly taking into account interference caused by the learning of intervening stimuli, which is a crucial characteristic of the Rescorla-Wagner rule.

Within the current study, we explore the effect of continuous learning with a model of the mental lexicon called the Discriminative Lexicon Model (DLM), and its learning mechanism, Linear Discriminative Learning (LDL). The DLM posits simple modality-specific mappings between numeric representations of words' forms and numeric representations of their meanings \citep{Baayen2018, baayen2019discriminative}. The DLM has been successful both in modelling different morphological systems across a range of languages, such as Latin, English, German, Estonian, Korean and Maltese \citep{Baayen2018, baayen2019discriminative, chuang2020estonian,chuang2021vector,heitmeier2021modeling, nieder2021comprehension}, but at the same time also at modelling a range of behavioural data \citep{cassani2019nonwords,chuang2020processing, heitmeier2020simulating, heitmeier2021modeling,schmitz2021durational, shafaei2021ldl,  stein2021morpho}. It implements learning using an error-driven learning rule for continuous data \citep{Widrow1960, milin2020keeping} which is closely related to the later developed Rescorla-Wagner rule. Additionally, in contrast to previous models such as Naive Discriminative Learning \citep[NDL; ][]{baayen2011amorphous}, it uses word embeddings to represent words' semantics. Word embeddings (aka semantic vectors) represent meanings in a distributed manner, building on the hypothesis that similar words occur in similar contexts \citep{Harris1954Distributional}. They are able to capture fine-grained meaning similarities between words and have been shown to predict numerous aspects of human processing in various studies \citep[e.g.][]{baayen2019discriminative, baroni2014don,mandera2017explaining, Westbury:Keith:Briesemeister:2014, westbury2022you}.

The computational modeling study that we report below is motivated by two hypotheses.  First, we anticipate that learning takes place not only in priming trials, but from trial to trial in unprimed tasks such as simple lexical decision, and by inference, in daily life, from word use to word use. Second, we take on the challenge of demonstrating that the DLM is powerful enough to predict the consequences of trial-to-trial learning for reaction times at the detailed level of individual subject-item combinations.
The current study therefore improves on previous modelling work in four ways: a) we will use an error-driven learning algorithm, building on previous work demonstrating its success in modelling a wide range of phenomena in psycholinguistics, b) we aim to model the learning task at a much more fine-grained level than previous work \citep[e.g.][]{jones2013sequential} by taking into account both words' forms and their meanings,  c) we will take into account a much larger set of stimuli presented in a much longer experiment compared to previous work \citep[e.g.][]{oppenheim2010dark}, and d), last but not least,  we will demonstrate that learning effects can be found (and predicted in fine detail using Linear Discriminative Learning) in experiments not specifically designed to detect learning effects.

Being a simple psycholinguistic experiment with a long history in the field, megastudies of lexical decision are now available, experiments which have recorded lexical decision data for large numbers of participants and for thousands of experimental stimuli for various languages, such as English, Dutch or Spanish \citep{aguasvivas2018spalex, balota2007english, brysbaert2016impact, keuleers2012british}. In the present work, we use data from the British Lexicon Project \citep[BLP;][]{keuleers2012british}, which encompasses lexical decision data from 78 participants for about 28,000 words and an equal number of nonwords. With datasets as big as these, even small effects of trial-to-trial learning should be detectable, if they exist.

In order to test our main hypothesis that during psycholinguistic experiments continuous learning occurs and can be traced down to fine-grained word-level updates of mappings between word forms and meanings as modelled by the DLM, we proceeded as follows. We first implemented two instances of the DLM to predict participants' lexical decision reaction times: one with learning updates of the lexicon after each trial and one without any learning updates.  We then tested which of these two models provides a better fit to reaction times. If the model with incremental updates shows better model fit, we can conclude that continuous learning may indeed be taking place during the experiment \citep[see][for a similar approach comparing models capturing inter-trial priming effects in a visual search task]{allenmark2021inter}.

In addition to this main question, we also explored two further issues. Firstly, we examined what the model tells us about lexical processing in general. The form and meaning representations and learning mechanisms that we are using in the present study have been found to be useful for predicting behavioural data in previous work \citep[e.g.][]{chuang2020processing, schmitz2021durational, stein2021morpho}, but for an improved understanding of what insights they offer, we compare the measures that we extracted from the DLM with classical psycholinguistic predictors such as orthographic neighbourhood density. Thus, we explore whether we still need such classical predictors or whether our model-based ones render them superfluous.

Secondly, we explore individual differences. Previous work has shown that there are considerable individual differences in
lexical processing. For example, \citet{kuperman2011individual} observed that in highly skilled readers, the frequency of the base word of morphologically complex words predicted longer reading latencies, whereas in low-skilled readers, it predicted shorter ones. Orthographic effects also differ across individuals. \citet{milin2017learning} conducted a serial reaction time experiment which they also simulated with NDL. They found that readers who speed up more across the experiment are less influenced by how much the target word is predicted by its orthographical cues than other subjects. Further studies confirm the influence of individual differences \citep[e.g.][]{fischer2018individual, perfetti2005word}, but note that connecting differences in morphological processing to individual psychological measures is not straightforward  \citep{loo2019effects}. In the present work we explore individual differences in lexical processing in considerable detail by investigating the random effect structure of a linear mixed model,
in the hope of being able to provide an algorithmic characterization of these differences.

The paper is structured as follows: Section~\ref{sec:background} gives an overview over previous computational models of lexical decision, and how the DLM relates to them. Section~\ref{sec:ldl} introduces the DLM and Section~\ref{sec:ld} explains how lexical decision is modelled in the framework of the DLM. In Section~\ref{sec:training} we give details on data pre-processing and the statistical models we employed to answer our main research questions. Section~\ref{sec:results} reports our findings regarding insights into lexical processing and the lexical decision task
which we can gain from the DLM, the effect of trial-to-trial learning as well as individual differences. Finally, Section~\ref{sec:discussion} discusses the conclusions which can be drawn from our results.

\section{Computational models of Lexical Decision}\label{sec:background}

There exists a multitude of models of word recognition and lexical decision, beginning from so-called ``box-and-arrow'' models, which describe the processing of stimuli only verbally, all the way to full-fledged computational models. The latter set of models has the advantage that they need to specify each aspect of the model precisely and that they can predict behavioural data quantitatively, resulting in models which can be tested rigorously \citep[e.g.][]{broker2020representing,dell2008introduction, luce1995four, mcclelland2009place}. This section gives a short overview of the most influential computational models which have been used to account for lexical decision, before contrasting them with the present approach.

\citet{norris2013models} classifies computational models of reading and word recognition into different ``styles'' such as interactive activation (IA), mathematical-computational, and connectionist models. IA models are essentially networks with typically three different feature levels: letter features, letters, and words, implemented as nodes in the network. Nodes typically inhibit other nodes at their own level, and activate or inhibit nodes at higher levels.  In order to recognise a word, first, relevant letter features are activated, which in turn activate letters which finally lead to activation of a word node fitting best to the activated letters \citep{rumelhart1982interactive}. Models based on the original IA model usually took this basic architecture for granted and refined single aspects \citep[``nested modelling'',][]{jacobs1994models}, such as the Spatial Coding Model \citep{davis2010spatial}, the Dual Route Cascaded Model \citep{coltheart2001drc} or the Multiple Read-Out model \citep{grainger1996orthographic}. IA models are commonly initialised by assigning resting activation levels to the individual nodes. For word nodes these can be derived from word frequencies \citep[Chapter 7]{mcclelland1989explorations}. The original versions of the three models mentioned here did not include an account of learning, but learning mechanisms were developed for some of the later iterations of these models \citep[e.g.][]{pritchard2016modelling}.

The second group, mathematical-computational models, are generally defined by mathematical functions rather than a network. The Diffusion Model \citep{ratcliff2004diffusion, wagenmakers2008diffusion} is such a model. The model takes frequency and type of nonword as given, and uses these to let the response drift slowly either to a word or nonword response, the aim being to account for the distribution of reaction times in lexical decision. The model's parameters are usually either set by the modeller or estimated from existing data. The Bayesian Reader \citep{norris2006bayesian} makes use of Bayes' formula to integrate the prior probability for various strings to be words (based on word frequency) with the incoming information on the target string to predict whether the string is a word or not.

A third style of models are so-called connectionist models. These models employ distributed representations rather than localist representations, and they usually make use of backpropagation of error \citep{rumelhart1986backprop} to estimate optimal connection weights.  The use of distributed representations makes it possible to model fine-grained meaning similarities and differences. One example of an influential connectionist model is the triangle model \citep{harm2004computing, seidenberg1989distributed}, which consists of orthography, phonology and semantic representations with mappings between them. The model can be trained, i.e. it ``learns'', and lexical decisions have been based on the error scores in these mappings \citep{seidenberg1989distributed}. The model was later implemented as a recurrent neural network to enable the modelling of reaction times based on time steps \citep{chang2013modelling}.

These models differ in their ability to (theoretically) implement trial-to-trial learning. For instance, while the original IA model does not include a learning mechanism, trial-to-trial effects could for example be implemented by not resetting activations after each trial (as described in \citeauthor{davis2006masked}, \citeyear{davis2006masked}, for primed lexical decision; see also discussion in \citeauthor{perea2003sequential}, \citeyear{perea2003sequential}). Both the diffusion model and the Bayesian Reader do not make explicit assumptions about the nature of the lexicon and instead only provide mechanisms for lexical decision-making itself. While an implementation of trial-to-trial adaptations in the decision process are imaginable \citep[see e.g.][for an implementation of inter-trial effects in a visual search task in a diffusion model]{allenmark2021inter}, they are not the focus of the current investigation. Similarly, the Multiple-Read Out Model could theoretically also accommodate trial-by-trial effects \citep[as discussed in][]{perea2003sequential}. All of these models address inter-trial effects at a very high level that does not take into account form or semantic similarity across trials. On the other hand, connectionist models based on backpropagation \citep[e.g.][]{chang2013modelling, seidenberg1989distributed} should be able to implement trial-to-trial learning in a similar manner to the one proposed in the current study. However, to our knowledge this has not been attempted so far, and thus it is not known whether the resulting trial-to-trial learning is flexible enough to match participant behaviour.

Lastly, a more recent style of modelling has emerged which \citet{norris2013models} calls symbolic/localist models: Naive Discriminative Learning \citep[NDL,][]{baayen2011amorphous}. NDL posits mappings between vector representations of form (for different modalities) and meaning; instead of using backpropagation it makes use of the simplest form of error-driven learning, the Rescorla-Wagner rule \citep{hoppe2022exploration,marsolek2008antipriming,ramscar2013error, Rescorla1972, Schultz:1998,  Trimmer:2012}, or the equilibrium equations of \citet{Danks2003} for the Rescorla-Wagner equations. The framework has been used to model both primed and unprimed lexical decision reaction times \citep{baayen2011amorphous,baayen2020modeling, milin2017discrimination}. \citet{milin2017discrimination} used an extension of the model where localist meaning representations are understood as pointers to distributed meaning representations. Properties of this second embedding network were found to also be highly predictive for lexical decision times \citep{baayen2016frequency}.

In a pilot study, \citet{chuang2021discriminative} used the incremental NDL model (without this extension to distributional semantics) to account for trial-to-trial learning effects in lexical decision data of one subject in the BLP, showing that NDL models which update connection weights after each trial show a better fit to speaker data than those without updates.

In the current study we explore a different implementation of discriminative learning by making use of the Discriminative Lexicon Model (DLM). Just as NDL, form units and semantic units are linked up without intervening hidden layers.  Unlike NDL, semantic representations are not localist but distributed. The use of distributed semantic representations is motivated by a range of studies that have pointed out the significance of semantics not only in lexical access and processing in general, but crucially also in the lexical decision task. Several studies found that variables related to a word's semantics, such as the semantic density of a word \citep{chuang2020processing,hendrix2021word}, its imageability \citep{balota2004visual}, its availability of meaning \citep{chumbley1984word} and how well its form predicts its meaning \citep{hendrix2021word, marelli2015semantic, marelli2018database} are predictive for reaction times in lexical decision.

In what follows, we use word embeddings as distributed representations of words' meanings.  Word embeddings (also known as semantic vectors) have been found  useful for predicting a remarkable number of phenomena in cognitive science in general \citep{gunther2019vector}, and lexical processing in particular  \citep[e.g.][]{cassani2019nonwords,chuang2020processing,chuang2021vector,Gahl2022thyme, heitmeier2020simulating,schmitz2021durational, stein2021morpho}. By replacing the localist representations of NDL (which formally can be represented by vectors of zeroes and ones, with ones representing which stems and morphological functions are present) with corpus-based word embeddings, it becomes possible to study the consequences for lexical processing of subtle similarities in meaning. For instance, plural semantics of nouns have recently been found to depend on the semantic class of the noun stem in English \citep{shafaei2022sense} and on case in languages such as Russian \citep{chuang:2022} and Finnish \citep{nikolaev:2022}. Such subtle dependencies in semantics are beyond what can be accomplished with the localist coding of NDL, and are also outside the scope of hand-crafted featural representations as used by, e.g.,  \citet{oppenheim2010dark}.

\section{Introduction to the Discriminative Lexicon Model}\label{sec:ldl}
\ \\
\noindent
The DLM is a theory of lexical processing that seeks to understand comprehension and production as mediated by modality-specific distributed representations of form and distributed semantic representations that are shared across modalities.  For auditory form representations derived from the speech signal, the reader is referred to \citet{shafaei2021ldl}. For details on how speech production is modeled, see \citet{baayen2019discriminative} and \citet{luo2021judiling}. Across modalities, the DLM sets up mappings between distributed form and meaning representations using the simplest possible networks, i.e., networks with an input layer, an output layer, and no hidden layer.  Mathematically, this amounts to using multivariate multiple regression to predict form from meaning, and meaning from form.

For the modeling of reading, word's orthographic forms need to be represented in a distributed way.
In this study, forms are represented by binary cue vectors coding the presence and absence of letter trigrams.\footnote{Many other representations are possible, such as features for orthographic input based on Histograms of Oriented Gradient features \citep{dalal2005histograms, linke2017baboons} \citep[further overview in][]{heitmeier2021modeling}.}
By way of example, consider the wordform \textit{aback}. As a first step, its set of unique trigrams is extracted (\texttt{\#ab, aba, bac, ack, ck\#}), with \texttt{\#} denoting word boundaries. In a second step, in a vector where each value stands for a possible trigram in the lexicon, the trigrams present in \textit{aback} are now coded with 1, and all others with 0. The resulting vector is stored as a row vector in a matrix $\mathbf{C}$ together with the form vectors of all other wordforms in the lexicon:
$$
\renewcommand{\kbldelim}{(}\renewcommand{\kbrdelim}{)}\mathbf{C} = \kbordermatrix{ & \mathtt{\#ab} & \mathtt{aba} & \mathtt{bac} & \mathtt{ack} & \mathtt{ck\#} & \mathtt{\#ba} & \mathtt{\#la} & ... & \mathtt{lac} \\
    \text{aback} & 1 & 1 & 1 & 1 & 1 & 0 & 0 & ... & 0 \\
    \text{back} & 0 & 0 & 1 & 1 & 1 & 1 & 0 & ... & 0 \\
    ... & ... & ... & ... & ... & ... & ... & ... & ... & ... \\
    \text{lack} & 0 & 0 & 0 & 1 & 1 & 0 & 1 & ... & 1 \\}.
$$
\ \\

For representing words' meanings, we made use of \textit{GloVe} embeddings \citep{pennington2014glove} that were visually grounded using the method of \citet{shahmohammadi2021learning}. We explored embeddings generated with \textit{Word2Vec} \citep{mikolov2013efficient} and its visually grounded counterpart. However, evaluation on the data of participant 1 of the British Lexicon Project indicated that grounded \textit{GloVe} vectors are the best choice.\footnote{
Visual grounding as proposed by \citet{shahmohammadi2021learning} aligns existing embeddings with information from images, without letting the grounded vectors deviate far from their original purely text-based embeddings. In this way, the vectors absorb some of the information available in images, but do not lose abstract information which is only available in text. The set of vectors found by \citet{shahmohammadi2021learning} to perform best on various benchmark tests in NLP,  such as lexical similarity \citep[e.g. on the MEN dataset,][]{bruni2014multimodal}, have a dimensionality of 1024, which we accordingly used in our simulations.
}
Words' semantic vectors are stored as row vectors in a matrix $\mathbf{S}$ (values in the following example are simulated):
$$
\renewcommand{\kbldelim}{(}\renewcommand{\kbrdelim}{)}\mathbf{S} = \kbordermatrix{ & S_1 & S_2 & S_3 & S_4 & S_5 & S_6 & S_7 & ... & S_n \\
    \text{aback} & -0.11 & 0.13 & -0.06 & -0.16 & -0.33 & 0.46 & 0.37 & ... & 0.13 \\
    \text{back} & 0.22 & 0.32 & -0.28 & -0.42 & -0.19 & 0.37 & -0.24 & ... & 0.01 \\
    ... & ... & ... & ... & ... & ... & ... & ... & ... & ... \\
    \text{lack} & -0.11 & 0.4 & -0.02 & -0.21 & -0.31 & -0.09 & 0.34 &...& -0.16 \\}.
$$
\ \\

\noindent For modeling comprehension, we use a mapping $\mathbf{F}$ that approximates $\mathbf{S}$ from $\mathbf{C}$. As the mapping is approximate, albeit optimal in the least squares sense, borrowing notation from statistics, we write
\begin{align}
    \hat{\mathbf{S}} = \mathbf{CF}.
\end{align}
For any individual wordform (represented as a binary vector $\mathbf{c}$), we can obtain its meaning (predicted semantic vector $\hat{\mathbf{s}}$) via
\begin{align}
    \hat{\mathbf{s}} = \mathbf{c} \cdot \mathbf{F}.
\end{align}

\noindent In the same way we can also model the initial stage of speech production as a mapping from a word's semantics to its form vector. This is achieved simply by a mapping in the opposite direction, so from $\mathbf{S}$ to $\mathbf{C}$, using a second mapping matrix $\mathbf{G}$. Again this mapping is approximate:
\begin{align}
    \hat{\mathbf{C}} = \mathbf{SG}.
\end{align}
$\mathbf{G}$ can now likewise be used to obtain a word's predicted form ($\hat{\mathbf{c}}$) from its meaning ($\mathbf{s}$):
\begin{align}
    \hat{\mathbf{c}} = \mathbf{s} \cdot \mathbf{G}.
\end{align}

\noindent There are two ways in which $\mathbf{F}$ and $\mathbf{G}$ can be computed. The first method makes use of the linear algebra underlying multivariate multiple regression (details on how the endstate-of-learning can be estimated efficiently can be found in \citeauthor{Baayen2018}, \citeyear{Baayen2018} and \citeauthor{luo2021judiling}, \citeyear{luo2021judiling}). The mapping matrices $\mathbf{F}$ and $\mathbf{G}$ can be thought of as the result of infinite experience with words' forms and meanings. We therefore characterize this method as estimating the ``endstate-of-learning''.  The mapping matrices at the endstate of learning are optimal, in the sense that they are learned as best as possible, given the limitations that come with the linear mappings of multivariate multiple regression (and, equivalently, the use of networks without hidden layers).

The second method learns the mappings incrementally. Mappings are updated each time a word is encountered. As expected, the mapping between a word's form and its meaning becomes more accurate the more often it is encountered.
Since we make use of distributed rather than localist semantic representations as in NDL, we replace the discrete learning rule of Rescorla and Wagner with the continuous rule of Widrow and Hoff \citep{Widrow1960}. Firstly, let's focus on word comprehension. When at time step $t$ a word $w_t$ is encountered, which has a wordform $\mathbf{c}_t$ and a meaning $\mathbf{s}_t$, the mapping from form to meaning is updated in a way which decreases the error between the predicted and the target semantics, making the learning ``error-driven''. In the following equation, $\eta$ represents the learning rate (the only hyperparameter of the mapping).
\begin{align}\label{eq:whF}
\mathbf{F}_{t+1} = \mathbf{F}_t + \mathbf{c}_t^T \cdot (\mathbf{s}_t - \hat{\mathbf{s}}_t) \cdot \eta.
\end{align}
Since the next time the same word is encountered, the mapping will be more accurate, we refer to this update step as ``strengthening''  the mapping.  It is worth noting that a higher learning rate $\eta$ implies not only that a form-meaning association is learned faster, but also that form-meaning associations which are not encountered are unlearned faster.

Secondly, for production we use the same algorithm to update the $\mathbf{G}$ matrix:
\begin{align}\label{eq:whG}
\mathbf{G}_{t+1} = \mathbf{G}_t + \mathbf{s}_t^T \cdot (\mathbf{c}_t - \hat{\mathbf{c}}_t) \cdot \eta.
\end{align}

In the full DLM model, production is followed by a second step: The result of mapping a semantic vector onto a form vector results in a vector that specifies, for all trigrams, how well these trigrams are supported by the semantic vector.  However, in order to actually produce a word, it has to be decided a) which trigrams have enough support to be included in the wordform that is to be articulated, and b) in which order the trigrams should be arranged for articulation. Since the trigrams are partially overlapping, they contain internal information about possible orderings. Various algorithms are available for generating candidates and selecting the optimal candidate for articulation, see, e.g., \citet{Baayen2018} and \citet{luo2021judiling}.  Evaluation of accuracy then reduces to comparing the selected word candidate with the target word form. As in this study, we only make use of the first step, i.e. calculating $\hat{\mathbf{C}}$ using the mapping matrix $\mathbf{G}$, and  these later steps in the production process do not play a role, in what follows, only the properties of the predicted form vector $\hat{\mathbf{c}}$ will be of interest.

\section{Modeling lexical decision making}\label{sec:ld}

\noindent
Similar to previous work both in discriminative learning models \citep{baayen2013sidestepping, milin2017discrimination} and also other computational models such as the interactive activation model of \citet{dijkstra2002architecture}, we view lexical decision as a two-step process. First, the incoming stimulus is processed by the lexical processing system. In our view (for which we present evidence below), this involves a comprehension mapping from form to semantics, followed by a production mapping from meaning to form \citep[following evidence for an `inner loop' in word recognition, see below and][]{chuang2020processing, liberman1985motor, skipper2017hearing, pulvermuller2006motor}. Importantly, the DLM highlights that these are not distinct cognitive processes but rather integrated components of the word recognition process. Next, a lexicality decision is made by distinct cognitive control processes \citep[as e.g. proposed by][]{Gurney:Prescott:Redgrave:2001, Redgrave:Prescott:Gurney:1999} which take as input ``data'' provided by lexical processing components. Instead of explicitly modelling the decision process, we will make use of statistical models to tease apart lexical processing measures and establish their individual contributions to the final decision. Note that this differs from some of the previous models of lexical decision which generally try to derive word/nonword decisions from the models directly \citep[e.g. the activation of a word-node in the interactive activation model,][]{rumelhart1982interactive}. We adopt this approach for two reasons: a) we think that lexical decisions are based on a wide variety of factors which cannot be simply captured by a single variable (this is confirmed by the diverse set of measures we find influencing the decision process below), and b) our focus in the present study lies on whether trial-to-trial learning effects arise in the course of the initial stage of lexical processing, and do not investigate trial-to-trial effects in the decision mechanism \citep[which previous studies have already explored, see for example][]{jones2013sequential}.

In this section, we first introduce how we think trial-to-trial learning takes place in the course of a lexical decision experiment, using the DLM to generate predictions for form and meaning vectors.  We then introduce a series of measures that we calculate from these vectors, including measures such as a word's semantic neighborhood. Importantly, the values of these  measures will depend on the learning history of the preceding trials.

Sections~\ref{sec:training} and \ref{sec:results} report how we have used these measures to predict the time it takes to execute lexicality decisions, using non-linear regression models fitted to the time series of reaction times in the British Lexicon Project.

\subsection{
Lexical processes
}\label{sec:ld_ldl}

\subsubsection{Prior knowledge}

\noindent
Participants come to a lexical decision experiment with fully developed knowledge of the words of their language. In order to approximate this prior lexical knowledge that participants bring to the experiment, we set up mappings between form and meaning for all the words that are encountered during the experiment.  As described above in Section~\ref{sec:ldl}, the DLM can learn words in two ways: using the linear algebra of multivariate multiple regression, resulting in endstate-of-learning mappings; or alternatively, using the learning rule of Widrow and Hoff, applied word token by word token. This learning rule is computationally demanding, and prohibitively so for training data with millions of word tokens. In the absence of properly chronologically ordered training data, we opted for initializing participants' lexicons using endstate-of-learning mappings. A  detailed discussion of the different options available for estimating mappings is available in \citet{Heitmeier:Chuang:Axen:Baayen:2023}.

Matrices $\mathbf{F}$ and $\mathbf{G}$ initialized with the endstate-of-learning calculated for the entire set of 28,456 words in the BLP for which semantic vectors were available (details in Section~\ref{sec:data}) resulted in an accuracy of 61\% for comprehension.
For 81\% of the words, the targeted semantic vector was among the five closest semantic neighbours. Accuracy for production was at 50\%; for 65\% of the 28,456 words, the targeted form was among the top 10 candidates. A possible reason for this relatively low production accuracy is the irregularities that abound in the English spelling system. Another possible reason is that in the present study, the mappings between form and meaning are constrained to be linear.

\subsubsection{Trial-to-trial learning: processing steps}

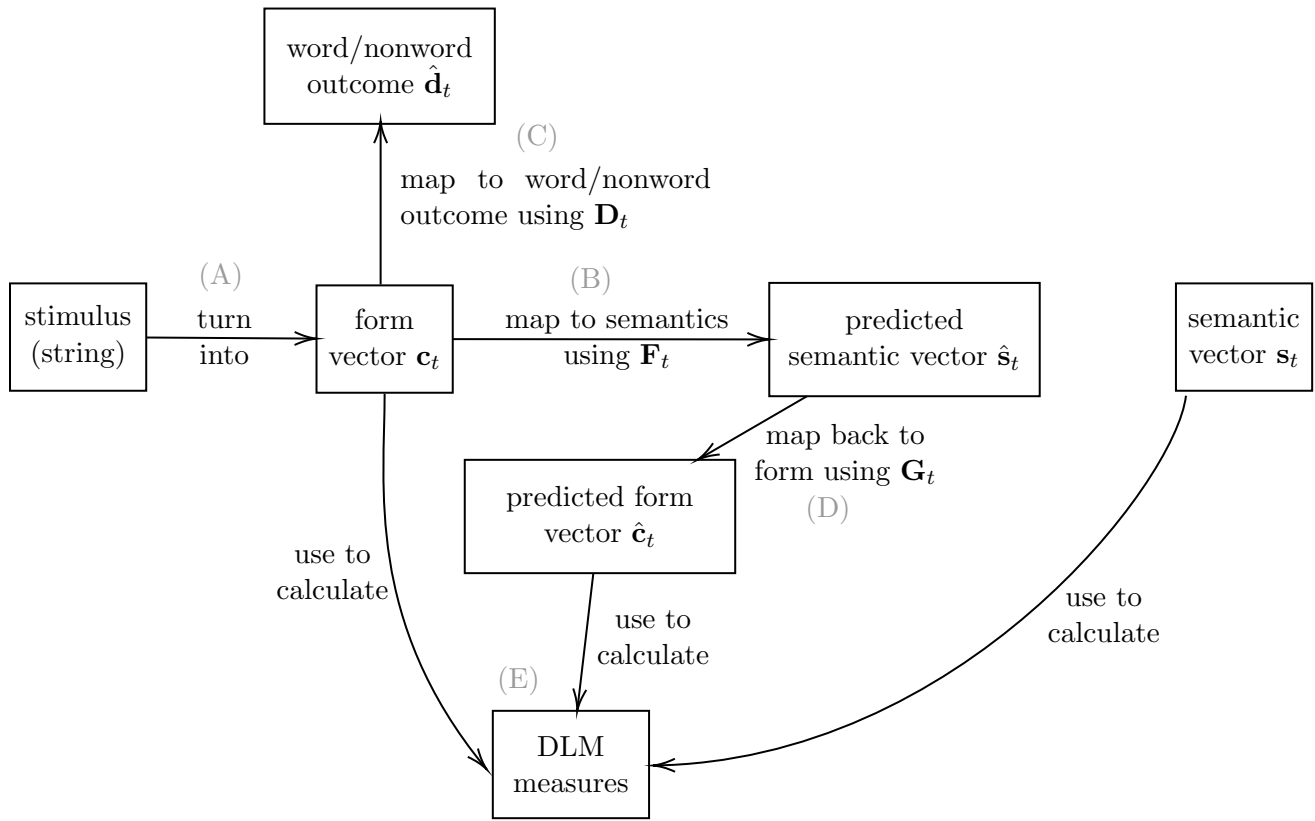
\begin{figure}[!ht]
    \centering
    \input{fig/ld_flowchart.tikz}
    \caption{Overview over steps during simulation of one trial $t$. Boxes represent representations, while arrows show processes. The last step (not pictured) is to update $\mathbf{F}_t$, $\mathbf{G}_t$ and $\mathbf{D}_t$ using the Widrow-Hoff learning rule.}
    \label{fig:trial}
\end{figure}

Having set up networks for participants' prior lexical knowledge, we now explain how we model a trial in the lexical decision experiment.  Figure~\ref{fig:trial} provides an overview of the different modeling steps that unfold at each subsequent trial.  When encountering a stimulus letter string at trial $t$, the very first step (labeled A in Figure~\ref{fig:trial}) is the encoding of this stimulus as a form vector $\mathbf{c}_t$. (Here and in what follows, we use a subscript $t$ to specify the state of a matrix or vector at trial $t$ in the experiment.) At this point, two processes are started up.  The first process (B) maps the form vector into the semantic space, using the comprehension mapping $\mathbf{F}_t$, resulting in the estimated semantic vector $\hat{\mathbf{s}}_t$ ($ \hat{\mathbf{s}}_t = \mathbf{c}_t \cdot \mathbf{F}_t$).

The second process (labelled C in Figure~\ref{fig:trial}) that is started up after the creation of the form vector is a mapping that learns to predict whether the form vector represents a word or a nonword. We assume that before the experiment, participants who have not participated in any lexical decision experiments before do not have experience with the meta-linguistic concept of `nonwords'.  Participants will know that there are words that they do not know the meaning of, and that words that they do know can be misspelled.  However, letter strings that are meaningless on purpose are not part of everyday language experience.  Readers who encounter a word they do not know are generally justified in assuming that the word is a meaningful part of their language, and they will seek  to infer its meaning from its context of use.  During the practice session preceding an actual lexical decision experiment, participants are therefore familiarized with the concept of nonwords, and we assume that this knowledge is subsequently developed and refined in the course of the experiment.\footnote{Our position differs from that of \citet{norris2006bayesian}, who argues that participants  make word/nonword decisions not only in the lexical decision task, but whenever they read. This claim seems to us especially unlikely in the light of recent results showing that even nonwords evoke semantics, see, e.g., \citet{cassani2020semantics, chuang2020processing}.}
The mapping from a form vector to a word/nonword outcome is formalized with a matrix $\mathbf{D}_t$. The support for the word/nonword outcome $d_t$ provided by the cue vector $\mathbf{c}_t$ given $\mathbf{D}_t$ is simply $d_t = \mathbf{c}_t \cdot \mathbf{D}_t$.
Note that this network does not represent a decision mechanism. Rather, we assume that the bottom-up support for word vs. nonword status is one source of evidence for the decision mechanism, which we take to be informed by other kinds of information as well, as explained below.\footnote{$\mathbf{D}$ is initialised with zeros at the beginning of the simulation. Ideally, it would be trained prior to the start of the simulation based on the training trials that each participant completed before the main experiment (see Section~\ref{sec:data}). Unfortunately, this part of the BLP data is not publicly available. Training $\mathbf{D}$ on the full BLP data would assume that participants have experience with making word/nonword responses prior to the experiment which we think is unlikely (see above).}

Recall that step B takes the form vector $\mathbf{c}_t$ and projects it into the semantic space, resulting in the predicted embedding $\hat{\mathbf{s}}_t$. We now introduce a `feedback loop' that takes the predicted embedding $\hat{\mathbf{s}}_t$, and projects it back into the form space, resulting in a form vector $\hat{\mathbf{c}}_t = \hat{\mathbf{s}}_t \cdot \mathbf{G}_t$. Evidence is accumulating that the comprehension and production systems interact and collaborate.  Multiple studies have reported empirical evidence that speech production is involved in speech perception \citep[e.g.][]{liberman1985motor,pulvermuller2006motor, skipper2017hearing}.  Feedback loops to production exist also during silent reading \citep[e.g.][]{haber1982does,abramson1997reader,perrone2012silent,kell2017phonetic,taitz2020motor}. Conversely, for speech production, \citet{levelt1983monitoring} proposed an inner loop from form to semantics \citep[see also][]{hartsuiker2001error}, and such a loop is implemented in the spiking neuron model of \citet{kroger2016modeling} as well as in the DLM \citep{baayen2019discriminative}.
More in general, \citet{casserly2010speech},  \citet{Hickok:2014}, and \citet{skipper2017hearing} argue for much better integration in linguistic and cognitive theories of the production and comprehension systems.
The feedback loop $\mathbf{G}$, which is assumed to be automatic and subconscious,  implements such an integration at a high-level of computational formalization.\footnote{
\citet{hickok2004dorsal} distinguish between dorsal and ventral streams in auditory comprehension, the former mapping sound to meaning, and the latter sound to articulatory-based representations. The dual pathway model allows for interaction between the two streams \citep[cf.][]{hickok2009functional}. Both mappings are represented in the DLM, which also has a mapping from meaning to articulatory representations, thus allowing the two streams to interact \citep[see][for detailed discussion]{chuang2020processing}. The DLM works with distinct, simple mappings, which guarantees a high degree of interpretability, but in the brain, the relevant networks are in all likelihood much more integrated and optimized. For a deep-learning model implementing more integrated (but also less straightforwardly interpretable) networks for comprehension and production, see \citet{schmidt-barbo_classification_2021}.
}
A feedback loop similar to the one proposed here was introduced in \citet{chuang2020processing}, and was shown to provide considerable leverage for predicting both naming latencies and spoken word duration in an auditory lexical decision task.

In the present study we model visual comprehension and therefore loop back to orthography. However, it remains an open question if a loop back to phonology might perform even better also for visual comprehension. We note here simply that linear mappings between orthography and phonology are generally quite accurate and the two could presumably be exchanged easily.

Once the predicted semantic and form vectors $\hat{\mathbf{s}}$ and $\hat{\mathbf{c}}$ have been obtained, the last step (E) is to  calculate various measures which will be used as predictors for reaction times in regression models. These measures will be introduced and discussed below in Section~\ref{sec:measures}.

\subsubsection{Trial-to-trial learning:  updating mappings}
Finally, we need to implement the learning which we hypothesise to take place after each trial. The participant's response is used to update all mappings (not displayed in Figure~\ref{fig:trial}). Using the Widrow-Hoff learning rule (see equations \ref{eq:whF} and \ref{eq:whG} above), the mapping $\mathbf{F}_t$ from cue vector $\mathbf{c}_t$ to its target semantic vector $\mathbf{s}_t$ is updated, as well as the mapping $\mathbf{G}_t$ from $\mathbf{s}_t$ to $\mathbf{c}_t$, both with learning rate $\eta = 0.001$, which we found to give best results for participant 1 in the BLP (see Section~\ref{sec:regression_modelling_strategies} for details on how hyperparameters were chosen). It is at this step that trial-to-trial effects arise.  If exactly the same stimulus would be presented again, the mapping to its semantics would be more accurate than before the update, resulting in `facilitation'. If a similar input stimulus with very different semantics would be presented next, the mapping would be less accurate. The cues that the target stimulus shares with the previous stimulus have just been mapped more strongly towards the meaning of the previous stimulus, resulting in `inhibition'.

The target semantic vector for updating $\mathbf{F}$ necessarily depends on the response of the participant and the lexicality of the stimulus. We distinguish four cases, as shown in Table~\ref{tab:reinforcing_semantics}. For word responses to words, the gold standard vector  generating the error is simply the semantic vector $\mathbf{s}_t$ of word $w_t$ in the semantic matrix $\mathbf{S}$.  The assumption here is that the participant understood the stimulus correctly, and that hence updating with $\mathbf{s}_t$ is justified.  We do not know this for sure, but it seems more likely that upon reading the word \texttt{dog}, some kind of dog came to participants' minds, rather than CO$_2$ or G\"{o}del's theorem. Occasionally, participants will have misunderstood the stimulus word
\citep[see also][]{diependaele2012noisy}, and although this certainly will add noise to our modeling efforts, this noise is unlikely to dominate results.

In trials where the participant responds with ``word'' but the stimulus is actually a nonword, we do not know which word the participant had in mind, or even whether the participant acted on a general sense that the stimulus was more word-like than non-word like.
We therefore assume that for this kind of trial, the error comes from a generalized sense of wordness.
To approximate this sense of wordness, we calculated the average of all word vectors in the participant's lexicon ---  the centroid of the cloud of word exemplars in the semantic space --- and we use this centroid to represent `wordness'.

For nonword responses, we need a semantic representation for what it means to be a nonword.  Without an embedding for `nonword', it is simply not possible to update mappings for trials with nonword responses.  We assume that a semantic representation for nonword does not exist before the experiment, but comes into being during the experiment. Dealing with nonwords is a metalinguistic skill that is acquired and continuously refined as the experiment proceeds.

An important property of the mapping $\mathbf{F}$ is that it generates semantic vectors not only for word stimuli, but also for nonword stimuli. The resulting nonword embeddings typically do not give rise to conscious percepts, but they do have detectable consequences for lexical processing \citep[see][for experimental evidence]{cassani2020semantics,chuang2020processing}.  Unfortunately, a nonword's predicted embedding $\hat{\mathbf{s}}_t$ cannot itself drive error feedback, as this error would be zero.  We therefore need an evolving nonword vector that reflects past experience with nonwords and their meanings.  We defined such a dynamic target semantic vector $\mathbf{n}_t$ for a nonword encountered at trial $t$ using the following recurrence equation:
\begin{align}
    \mathbf{n}_{t} = \frac{\mathbf{n}_{t-1} + \hat{\mathbf{s}}_{t-1}}{2}.
\end{align}
For trials in which the participant provides a word response, $\mathbf{n}_t$ does not change.
Thus, the current target nonword embedding is the average of the previous nonword embedding and the semantic vector generated from the previous nonword stimulus.\footnote{ This recurrence equation was developed using the data of participants 1 and 2 of the BLP.  The reader is referred to the Supplementary Materials for alternative solutions for calculating nonword embeddings.}
This implies that the embedding for the meaning of `nonword' is to 50\% determined by the last stimulus with a nonword response, with the nonword encountered before that (according to the participant's response) contributing 25\% to the vector, and so on. As a consequence, the nonword vector fluctuates considerably across the course of the experiment, with the magnitude of change determined primarily by the nonword and its estimated semantic vector encountered previously. Such a representation worked best for our validation subjects (see Section~\ref{sec:regression_modelling_strategies} below) and is in line with findings that category judgments show a recency effect with both a decisional and perceptual component (\citeauthor{jones2006recency}, \citeyear{jones2006recency}, but see \citeauthor{duffy2008primacy}, \citeyear{duffy2008primacy}, for a possible primacy effect in category induction).

\begin{table}[]
    \centering
    \begin{tabular}{c|c|c}
    \hline
        Lexicality & Response = Word & Response = Nonword \\
        \hline
        Word & reinforce using word's semantic vector & reinforce using nonword vector\\
        Nonword & reinforce using average of all semantic vectors & reinforce using nonword vector \\
        \hline
    \end{tabular}
    \caption{Decision table of which vector is chosen as target semantic vector for updating $\mathbf{F}$ after a trial.}
    \label{tab:reinforcing_semantics}
\end{table}

We now have in place all vectors required for updating the mappings $\mathbf{F}_t$ and $\mathbf{G}_t$.  What remains to be clarified is how the mapping $\mathbf{D}_t$ from form to word/nonword outcome is updated from trial to trial.
We update the mapping matrix $\mathbf{D}_t$ with the Widrow-Hoff learning rule, the target outcome being the participant's word/nonword response $r_t \in \{1, 0\}$. Crucially, $\mathbf{D}_t$ is not updated according to the actual lexicality of the stimulus, but strictly according to the participant's response. Since there is no ``correct/incorrect'' feedback in the BLP,  we are constrained to modeling the participant's individual experience of the experiment. Therefore,
\begin{align}
    \mathbf{D}_{t+1} = \mathbf{D}_t + \mathbf{c}_t^T \cdot (r_t - d_t) \cdot \eta,
\end{align}
with $d_t = \mathbf{c}_t \cdot \mathbf{D}_t$.

\subsubsection{Trial-to-trial learning: learning rates}

\noindent
Based on exploration with the data of subject 1, the learning rate $\eta$ was set to 0.01 for mapping $\mathbf{D}$, and to 0.001 for the mappings $\mathbf{F}$ and $\mathbf{G}$. It makes sense that the learning rate for the word/nonword outcome is an order of magnitude higher than the learning rate used to reinforce the mappings between forms and meanings. The lexical decision task requires subjects to make metalinguistic judgements in a cognitive task that subjects do not have much experience with, and that they learn to rapidly optimize as the experiment unfolds \citep{baayen2022note}. By contrast, lexical knowledge in long-term memory is expected to be much less affected by trial-to-trial contingencies.

In what follows, we used the same learning rates $\eta = 0.001$ for $\mathbf{F}$ and $\mathbf{G}$, and $\eta = 0.01$ for $\mathbf{D}$ for all participants. The assumption that learning rates are fixed across participants involves substantial simplification, but it protects us from having to solve an extremely complex high-dimensional optimization problem.

\subsection{Predicting reaction times}\label{sec:measures}

\noindent
For assessing whether incremental learning in the course of the experiment is taking place, we make use of generalized additive regression models (GAMs) fitted to participants'  response latencies\footnote{
According to \citet{thul2021using}, GAMs are complex, advanced techniques that are not fully understood and that come with potential side-effects. However, \citet{baayen2022note} show that the problem reported by Thul et al. is due to a bug in the \textbf{mgcv} package, which has been fixed from version 1.8-36 onwards.
}\footnote{We also ran two generalised linear mixed models predicting participants' decisions from the same set of predictors. These models gave very similar results to the models based on reaction times and can be found in the Supplementary Materials.}.  We distinguish between two kinds of predictors: classical predictors with a long history of exploration, and model-based predictors. The former are invariant with respect to experimental time (trial), the latter crucially depend on the learning history in the course of the experiment.  We discuss these predictors in turn.

\subsubsection{Classical predictors}
Three psycholinguistic, non-incremental predictors have been used many times to predict lexical decision reaction times \citep[e.g.][]{balota2004visual, keuleers2012british, yap2015responding}.

\textbf{Word Frequency}, i.e., the frequency of occurrence of a word in some corpus, is generally associated with shorter reaction times in lexical decision tasks \citep[e.g.][]{keuleers2012british, rubenstein1970homographic, scarborough1977frequency}. We used word frequency counts based on the British National Corpus\footnote{\url{http://www.natcorp.ox.ac.uk}}, as reported in the BLP data. Though subtitle frequencies have been reported to be superior at predicting reaction times \citep{brysbaert2009moving}, we opted for frequencies from the BNC because, first, this corpus covers all registers and second, the confound of frequency and arousal found in subtitle corpora \citep[cf.][]{baayen2016frequency} is avoided.

\textbf{Word length}, measured in terms of number of letters, is a predictor the effect of which is still under debate \citep[overview in][]{new2006reexamining}. Null effects reported for this predictor may have arisen from a failure to match word and nonword stimuli in lexical decision experiments, see \citet{chumbley1984word}. Word length has also been reported to have a U-shaped effect on reaction times  \citep{baayen2005data, new2006reexamining}. The latter study reports that in the English Lexicon Project \citep{balota2007english}, word lengths up to 5 letters tend to give rise to shorter reaction times, and lengths from 8 to 13 letters to longer reaction times. No effect was found for lengths between 5 and 8 letters. \citet{hendrix2021word}, using survival analysis, found that the effect of word length changes across the distribution of reaction times. Early responses are unlikely for long words, presumably because of higher visual processing costs linked to longer words. For short words, early responses are much more likely. Later responses are somewhat more likely for longer words. However,  very late responses appear to be equally likely for all word lengths. For nonwords, on the other hand, multiple studies found that word length elicits longer reaction times \citep{balota2004visual,yap2015responding}.

\textbf{Orthograhpic Neighbourhood Size} has been reported to afford shorter reaction times for words \citep[see, e.g.,][]{andrews1992frequency,balota2004visual}. On the other hand, orthographic neighbourhood size was not found to be predictive for reaction times to words in various virtual experiments, where reaction times for stimuli used in other studies were retrieved from the BLP \citep{keuleers2012british}. For nonwords, \citet{yap2015responding} and \citet{balota2004visual} observed that larger neighbourhood size led to longer reaction times.

Similar to \texttt{Word Length}, the effect of \texttt{Orthographic Neighbourhood Size} thus seems to be somewhat unclear with regard to words, but clearly leads to longer reaction times for nonwords. In the analyses reported below, we quantified orthographic neighbourhood size by the number of words in CELEX \citep{baayen1995celex} with a Levenshtein distance \citep{levenshtein1966binary} of 1 from the target stimulus.

In our analyses, we also included two task-related predictors. \textbf{Trial Number} denotes the rank of a stimulus in the experimental list. The reaction times in a lexical decision experiment constitute time series, and these time series often show structure, indicating that the responses are not independent.
\texttt{Trial Number} gauges three distinct processes that often unfold in the course of experiments.  First, for most of the participants, reaction times decrease substantially as \texttt{Trial Number} increases. In the BLP, participants adapt to the task and generally respond more quickly as the experiment proceeds \citep{keuleers2012british}.
We interpret this as reflecting participants tuning in to the lexical decision task.  Explaining this kind of learning process  is outside the scope of the present study, which focuses on lexical learning and not on how participants optimize task behavior. Second, in the course of an experiment, many participants reveal fairly large ups and downs in response times that show up as undulating, wave-like patterns in plots of reaction time against \texttt{Trial Number} \citep[see, e.g.,][]{baayen2017cave}.  Such variable behavior appears to be more pronounced for participants with higher degrees of ADHD \citep{baayen2022note}. Undulations in response behavior most likely reflect fluctuations in attention.   Third, it cannot be ruled out that \texttt{Trial Number} also captures, in part, the much more modest consequences of ongoing low-level lexical learning and recalibration.

We included \texttt{Trial Number} as predictor in our GAM models, which offer powerful tools for capturing nonlinear effects, in order to bring the large variances that are due to learning and changes in attention under control. By doing so, when testing models with measures gauging incermental lexical learning, we work against our hypothesis, as effects of lexical learning could be absorbed by the effect of Trial Number.

\textbf{Response Type} We also included the participant's response (word/nonword) as a binary predictor.   Responses to words and nonwords tend to differ systematically \citep{keuleers2012british}, depending on the kind of nonwords used \citep{ratcliff2004diffusion}.  Since both correct and incorrect responses are an integral part of the learning process, we included both types of responses in our analyses, adding a factorial predictor to differentiate between response types. An additional reason for including response as a predictor is the following: given that different target semantic vectors are used depending on whether a participant's response was `word' or `nonword', we reasoned that it is possible that a DLM-based measure is significant due to a confound with response type. We controlled for this potential confound by adding response type as an additional predictor.

\subsubsection{Measures from the DLM}
From the DLM, we derived five measures for predicting the reaction times in the BLP. Our method for selecting these measures is described in Section~\ref{sec:regression_modelling_strategies} (see the Supplementary Materials\footnote{Supplementary Materials including the simulation code, all generated measures and statistical analyses can be found at \url{https://osf.io/bxmt2/}.} for a full listing of all measures that we investigated).

The first measure assesses words' \textbf{Semantic Density}, the number and proximity of its closest semantic neighbors.  Measures of semantic density have been used in previous work to predict not only reaction times in lexical decision \citep[e.g.][]{buchanan2001characterizing,chuang2020processing,hendrix2021word,  schmitz2021durational, stein2021morpho}, but also in other fields such as word learning \citep{hopman2018predictors}. The measure of semantic density that we have found to be optimal is based on the closest semantic neighbors of the predicted semantic vector $\hat{\mathbf{s}}$, and gauges how densely populated the area in semantic space is around $\hat{\mathbf{s}}$.  If a form vector $\mathbf{c}$ is projected by the mapping $\mathbf{F}$ into a semantically dense area, this indicates not only that the predicted vector $\hat{\mathbf{s}}$ has landed in an area of high lexicality, providing it with a high degree of ``wordlikeness'', but also that it might be more difficult to tell the meaning $\hat{\mathbf{s}}$ of the word apart from similar meanings \citep{arnold2017words}.

Semantic density can be quantified by inspecting the $n$ closest semantic neighbours and computing the mean of their cosine similarities to $\hat{\mathbf{s}}$ \citep[see e.g.][]{buchanan2001characterizing}. Let $CS_t$ be the set of all cosine similarities between $\hat{\mathbf{s}}_t$ and the semantic vectors $\mathbf{s}_k \in \mathbf{S}$:
\begin{align}
        CS_t = \{\text{cosine\_similarity}(\hat{\mathbf{s}}_t, \mathbf{s}_k) \, \forall \, \mathbf{s}_k \in \mathbf{S}\}.
\end{align}
Then, \texttt{Semantic Density} is defined as the mean of the $n$ highest values in $CS_t$:
\begin{align}
         \text{Semantic Density}_t = \frac{\sum \text{max}_{n}(CS_t)}{n}.
\end{align}
We set $n=10$.

A second semantic measure, \textbf{Form-driven Semantic Relatedness}, assesses how close the semantic vectors are of a word's orthographic neighbors.  This measure is  motivated by two findings from previous work. Firstly, we know from studies such as \citet{bowers2005automatic,forster1984repetition, rodd2004leotards} that during word recognition, the meanings of orthographic neighbors are activated.
Secondly, \citet{marelli2015semantic} proposed a measure of the semantic similarity between embeddings of word's orthographic neighbours (Orthographic-Semantic Consistency, OSC), and reported that it is predictive for lexical decision latencies in the BLP.  \texttt{Form-driven Semantic Relatedness} follows up on these findings by quantifying how far apart the embeddings of orthographic neighbours of a stimulus are in the semantic space.

\begin{figure}
    \centering
 \input{fig/shortest_path_diagram.tikz}
    \caption{Four points in a two-dimensional semantic space, with hypothetical (euclidean) distances between them. The green node is the vector $\hat{\mathbf{s}}$ for the target word \textit{back}, the others represent the semantic vectors of four of its orthographical neighbours. \texttt{Form-driven Semantic Relatedness} measures the shortest path connecting all points. In this toy example, the shortest path would be $back \rightarrow lack \rightarrow tack \rightarrow sack \rightarrow back$, with a length of 9. The \texttt{Form-driven Semantic Relatedness} for this example therefore is 9.}
    \label{fig:shortest_path}
\end{figure}
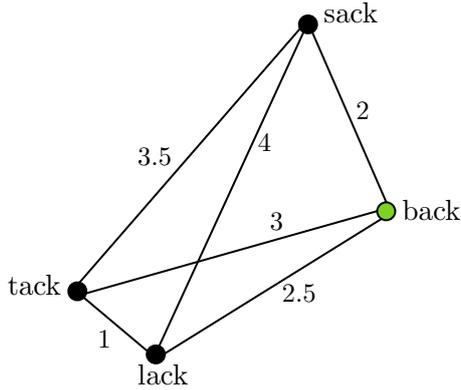

Let $N$ denote the set of a word's nearest orthographic neighbours, defined as all words with the same number of letters, and one letter exchanged, following \citet{Coltheart:77}. We calculate the corresponding predicted semantic vectors $\hat{\mathbf{s}}_n$ for each neighbor $n \in N$. Then we find the Form-driven Semantic Relatedness in the semantic space (measured in Euclidean distance) that connects all predicted semantic vectors $\hat{\mathbf{s}}_n$ including the predicted semantic vector of the target stimulus $\hat{\mathbf{s}}_t$ (see Figure~\ref{fig:shortest_path}).\footnote{
Finding the Form-driven Semantic Relatedness is a case of the \textit{Travelling Salesman Problem}, where the goal is to find the shortest path connecting all points in a multi-dimensional space. We made use of algorithms by \citet{pferschy2017generating}, implemented in Julia (\url{https://github.com/ericphanson/TravelingSalesmanExact.jl}).}  The Form-driven Semantic Relatedness measure is correlated with, but not identical to the OSC measure.  For the 54\% of words in the BLP for which OSC is available in \citet{marelli2018database}, the correlation between \texttt{Log Form-driven Semantic Relatedness} and OSC is $r=-.34$. OSC is a frequency-weighted average of \textit{cosine similarities}, whereas the \texttt{Form-driven Semantic Relatedness} measure evaluates the \textit{distances} between neighbor's embeddings; evaluation using cosine similarities in semantic space (rather than distances) is implemented in our \texttt{Semantic Density} measure. Important from a geometric perspective is that the combination of \texttt{Form-driven Semantic Relatedness} and \texttt{Semantic Density} allows us to probe semantic space both using angles and distances between semantic vectors.

These two semantic measures are complemented with two measures that evaluate the predicted form vectors generated in the ``feedback loop''.  Recall that the feedback loop uses the production mapping $\mathbf{G}$ to project a stimulus' predicted semantic vector $\hat{\mathbf{s}}_t$ back into the form space, resulting in the predicted form vector $\hat{\mathbf{c}}$.  \textbf{ C-Precision} measures how well the predicted form vector $\hat{\mathbf{c}}_t$ matches the original form vector $\mathbf{c}_t$, and is defined as the correlation between the two:
\begin{align}
        \text{C-Precision}_t = \text{cor}(\mathbf{c}_t, \mathbf{\hat{c}}_t).
\end{align}
With this measure, we probe whether the meaning that is understood maps back properly onto the corresponding form. We also evaluated the quality of $\hat{\mathbf{c}}$ with a second measure,
\textbf{Cue Activation Diversity}, the L1-norm of the predicted form vector:
\begin{align}
        \text{Cue Activation Diversity}_t = \sum_{j=1}^n |\hat{c}_j| = L_1(\hat{\mathbf{c}}_t),
\end{align}
with $n$ the length of \chat. This measure quantifies the uncertainty in the predicted form vector \chat \, \citep[similar to the activation diversity measure used in][]{milin2017discrimination}.\footnote{The L1 Norm of a vector measures the sum of the absolute values in the vector. In general, it will therefore be higher the more high values there are in the vector. If a vector is predicted correctly, there will be typically only a few values around one and most will be close to zero. However, in reality the average Cue Activation Diversity (not log-transformed) is 26.3 for words and 32.7 for nonwords in our dataset. This suggests two things: a) many more cues than the ones which actually occur in the stimulus are at least to some extent activated. This means that higher values likely indicate support for a range of different cues, which results in uncertainty. And b), Cue Activation Diversity is higher for nonwords than for words, which further supports our interpretation of this variable as ``uncertainty''.}

The last measure, \textbf{ Yes-activation}, assesses the ``wordlikeness'' of a word form, and is defined as the support for the outcome ``Word'' (the value of $d_t$, see Section~\ref{sec:ld_ldl}).  It thus measures how strongly the sublexical cues of the visual stimulus support a word outcome given the participant's previous experience with words and nonwords.

The four lexical measures (\texttt{Semantic Density}, \texttt{Form-driven Semantic Relatedness}, \texttt{C-Precision}, and \texttt{Cue Activation Diversity}) can be computed in two ways. They can be calculated for `dynamic simulations', i.e., simulations in which the mappings are updated after each trial, and as a consequence, vary from trial to trial.  Alternatively, in simulations without learning, they can be calculated on the basis of the mappings representing subjects' prior knowledge. In these static simulations, these measures always have the same values for a given word, irrespective of the participant and the moment in the experiment at which it is presented.
Of course, \texttt{Yes-Activation}, by its very nature, is available only for dynamic simulations.

%It is derived from the network $\mathbf{D}$ introduced specifically for the modeling of lexical decision making. Therefore, it is not directly related to lexical processing itself, but rather a task-specific, learned measure. It is therefore not included in the set of \textit{lexical-DLM} measures.

\section{Data preprocessing and regression modeling strategies}\label{sec:training}

\noindent
This section describes data preprocessing, and also provides details on our regression modelling strategies.

\subsection{Data}\label{sec:data}

\noindent
We used the data collected by \citet{keuleers2012british} in the British Lexicon Project (BLP). They collected lexical decision reaction times for 28,730 mono- and disyllabic words and an equal number of nonwords from 78 British students. To save time --- the experiment took about 16 hours per participant ---, each participant responded to half of the target stimuli. Words with a frequency of at least 0.02 per million in the BNC were selected. The nonwords were generated from real words (the `base' words) using Wuggy \citep{keuleers2010wuggy}, implementing the following constraints: (1) nonwords and words were matched in syllabic and subsyllabic as well as in morphological structure, (2) monosyllabic nonwords differed in one and disyllabic ones in two subsyllabic elements from the base word, (3) transition frequencies of subsyllabic elements were matched as much as possible. As described in previous work, even though all nonwords were based on real words, the method used to generate them made most nonwords opaque as to their base words \citep{hendrix2021word}.

Participants first completed a set of 200 training trials with trisyllabic words and matching nonwords to familiarise themselves with the task. Then, participants were allowed to freely choose how many blocks (500 trials) they wanted to complete in one day. There was no time-limit on responses, and no feedback was given during the experiment. Further details on the experimental procedure can be found in \citet{keuleers2012british}.

Selecting all words in the BLP for which a visually grounded \textit{GloVe} embedding \citep{shahmohammadi2021learning} is available resulted in a set of 28,465 words. Before the simulation, we removed trials with `null' and `nan' as target stimuli (156 datapoints), as these spellings disrupted data processing. We also removed all trials with time-out responses, as for these trials  (21 responses for subject 65, 4 for subject 70 and 1 for subject 10) no clear word/nonword response is available.  Finally, we excluded all trials with reaction times $\leq 100$ ms, which is the minimum for response execution, or $> 2000$ ms, which are outliers in the distribution and probably reflect  additional cognitive processes which are not of interest to the present study (20,094 datapoints, 0.9\% of the total dataset)).

The distribution of reaction times in the BLP has a strong right skew.  In order to make the reaction times suitable for analysis with Gaussian regression modeling \citep{ratcliff1993methods}, they were transformed as follows:
\begin{align}
    \text{RTinv} = -1000/\text{RT}.
\end{align}
The distribution of \texttt{RTinv} is close to normal.  This transformation implies that instead of response time, we model response rate (with a scaling factor 1000 to avoid very small numbers, and negative sign to ensure a positive correlation of the rate variable with the time variable). However, since a higher \texttt{RTinv} (i.e. lower response rate) corresponds to higher raw reaction times, for ease of exposition we will refer to this negative response rate as ``reaction time'' for the remainder of this paper.

\begin{figure}
\centering
\begin{subfigure}[b]{\textwidth}
\centering
    \includegraphics[width=\textwidth]{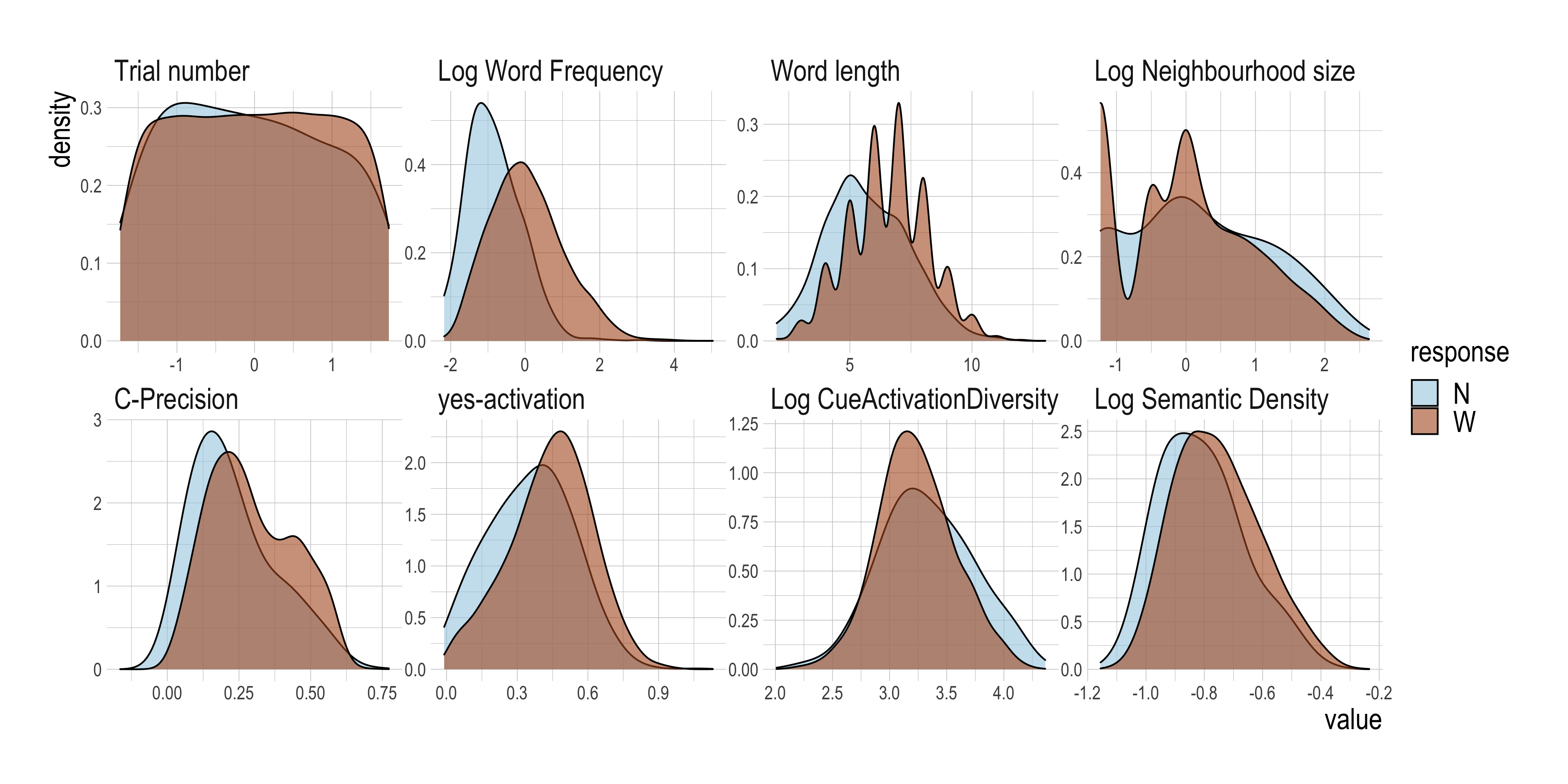}
    \caption{Words}
    \label{fig:density_words}
\end{subfigure}\hfill
\begin{subfigure}[b]{\textwidth}
\centering
    \includegraphics[width=\textwidth]{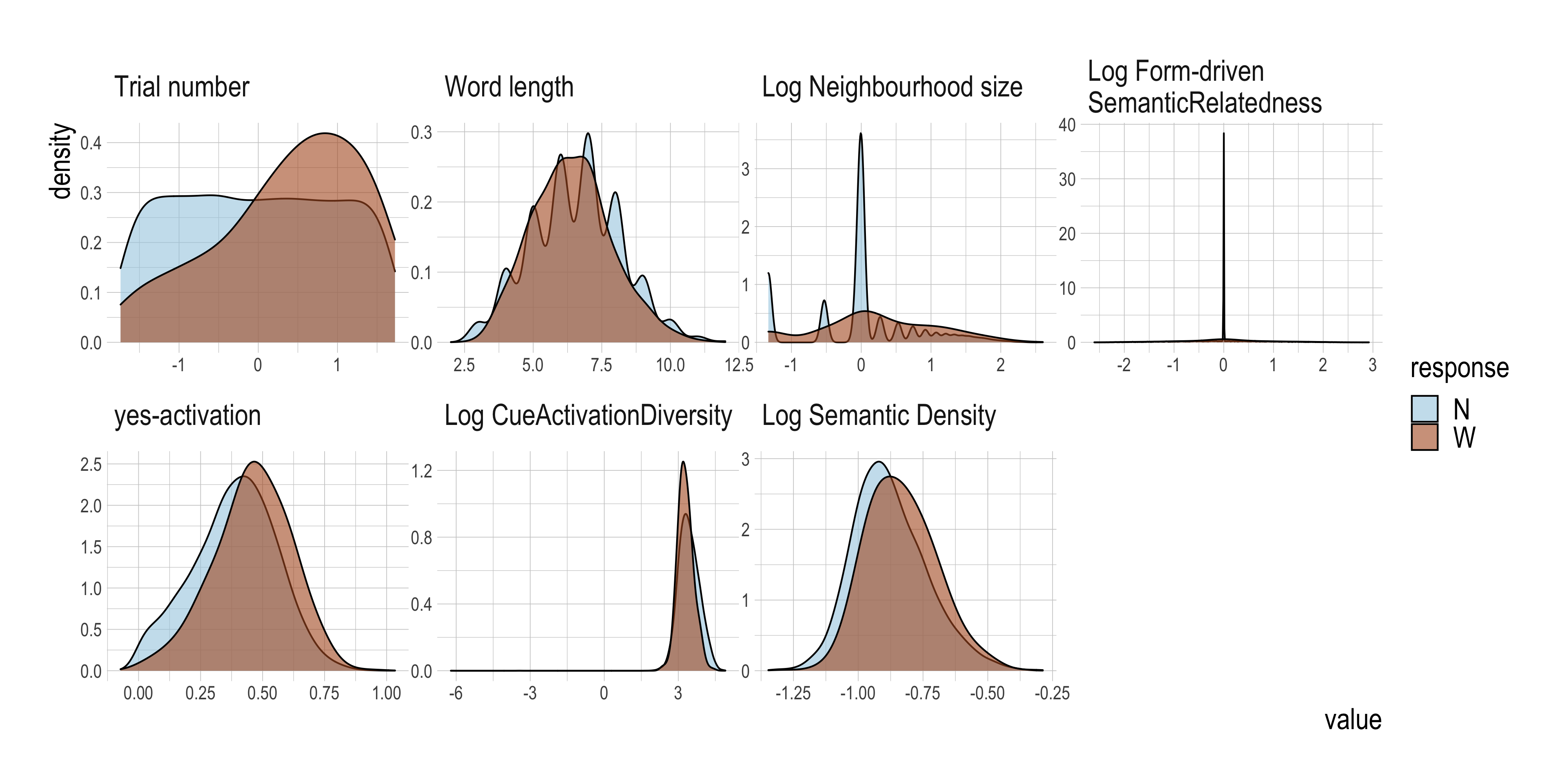}
    \caption{Nonwords}
    \label{fig:density_nonwords}
\end{subfigure}
\caption{Distribution of measures for words and nonwords for subject 1. }
\label{fig:density}
\end{figure}

For each predictor, we inspected its distribution. If this distribution showed a strong right skew with outliers, a log-transformation was applied (if necessary to back off from zero, 0.002 was added before taking logs). Figure~\ref{fig:density} presents the estimated probability density curves for words (upper panels) and nonwords (lower panels), based on the data of subject 1.

Special care was taken for predictors with a substantial number of zeros.   For such predictors, a log transformation often leads to a bimodal distribution. In Figure~\ref{fig:density_words}, such a bimodal distribution is visible for \texttt{Log Neighbourhood Size}.  For such a variable $p$, we introduced an indicator variable $b$ indicating where the (untransformed) variable is zero (i.e. a factor which is zero when untransformed $p$ is zero, and is one otherwise), and added $b + b\times p$ to the regression model. In this way, we capture the mean difference in \texttt{RTinv} for the zero and non-zero values of $p$, and at the same time enable the regression model to capture the relative contributions of the non-zero values of $p$.
This procedure was necessary for  \texttt{Log Word Frequency} (binary predictor \texttt{in\_bnc}), \texttt{Log Neighbourhood Size} (binary predictor \texttt{has\_neighbours}) and \texttt{Log Form-driven Semantic Relatedness} (binary predictor \texttt{has\_neighbours\_path}). This had the added benefit of removing the spike at 0 in the distributions of \texttt{Log Form-driven Semantic Relatedness} and \texttt{Log Neighbourhood Size}, resulting in their effects remaining interpretable in the regression models below. \texttt{Trial number} was centered and scaled.

\subsection{Regression Modeling Strategies}\label{sec:regression_modelling_strategies}

Predicting the response latencies of the participants in the BLP as well as possible, faces many challenges. This task requires solving a highly complex optimization problem that is beset by a range of problems.

First, there are many potentially relevant predictors: classical predictors, model-based predictors, and task-related predictors, as outlined above. As many of these predictors are correlated, regression modeling carried out with the aim of understanding how individual predictors co-determine the response variable is not served well by including all variables jointly, due to issues of collinearity and concurvity.\footnote{Concurvity, the counterpart of collinearity in the strictly linear model, estimates the extent to which the partial effect of a given predictor can be accounted for by the other predictors in the model.  When concurvity is high, it is unclear whether predictors with high concurvity scores make an independent contribution  to the model fit.  For discussions of collinearity and concurvity, see \citet{friedman2005graphical} and \citet{tomaschek2018strategies}.}  In order to safeguard the interpretability of our regression models, we decided to limit as much as possible the number of predictors that we took into consideration.

Second, predictors may have non-linear effects, and may enter into non-linear interactions.  To constrain the search space of regression models, we decided not to consider many of the different non-linear interactions that could be considered.

Third, predictors are not necessarily equally relevant for individual participants.  In principle, learning rates may vary from participant to participant, resulting in different sets of model-based predictors, one for each participant.  Furthermore, a predictor that is highly relevant for one participant may be irrelevant for other participants.  As determining optimal learning rates for all participants individually has an unjustifiably high carbon footprint, we used the same learning rates across all participants.  However, we did carefully monitor for how the effects of predictors varied with participant, and will report on our findings in detail below.

For clarification, our aim is not to provide globally optimized participant-specific models that best predict response latencies.  We have a more modest aim, namely, to show that trial-to-trial learning indeed takes place, and that this trial-to-trial learning can be approximated by our implementation of the DLM model. This simpler goal motivates the simplifying strategies described above.

\subsubsection{Model development strategy}

In order to determine reasonable learning rates, and to select a well-motivated subset of predictors, we followed a development strategy widely used in machine learning.  When developing a model, the available data are often partitioned into training data, validation data, and test data.  The model is trained on, unsurprisingly, the training data, hyperparameters and modelling decisions are based on the validation data (usually a small proportion of the available data), and then its performance is tested on the held-out test data.

For our purposes, the training data are the total set of words in the BLP from which we estimate the prior lexical knowledge for the model.  Here, we don't have any hyperparameters. Given the set of words, the mappings are completely determined.

As validation data, we used the data of participants 1 and 2,  which together cover all words and nonwords occurring in the BLP (see Section~\ref{sec:data}). We used the data from participant 1 to estimate the two hyperparameters of the model, the learning rate for the lexical mappings ($\eta = 0.001$) and the learning rate for predicting word/nonword status ($\eta=0.1$), as explained above.  Furthermore, we used the validation data to trim down the set of possible model-based predictors to a much smaller set of well-supported predictors, as detailed below.

The remaining 76 subjects constitute the test data on which we evaluate the combination of the prior lexical knowledge, the learning rates, and the selected predictors.  In this way, we make sure that we evaluate our computational model on data on which it has not been developed and fine-tuned \citep[see also][]{wilson2019ten, shmueli2010explain}.

\subsubsection{Variable selection} \label{sec:pred_selec}

As mentioned above, given a large number of predictor variables, many of which are to some extent correlated (the maximum correlation of a pair of DLM-based predictors was $r < .6$), in order to safeguard interpretability of the partial effects of predictors in our regression models, it is crucial to bring down the number of predictors.  For the full list of model-based predictors, the reader is referred to the supplementary materials.

Predictors were included in our exploratory models if, and only if, (1) their partial effect was significant ($p < 0.001$), (2) including the predictor improved the overall Akaike Information Criterion \citep[AIC;][]{akaike1998information}\footnote{AIC measures model fit while punishing model complexity. AIC makes it possible to compare the fits of two models: the bigger the difference in two AIC values, the more likely one model is than the other, given the data (smaller AIC values mean better model fit). By way of example, if model A has an AIC which is 100 points lower than that of model B, then model A is $e^{\frac{100}{2}} = 5.18 \times 10^{21}$ times more likely than model B.}, and (3) inclusion of a predictor did not lead to unacceptably high concurvity. We allowed for two exceptions to these rules: \texttt{C-Precision} in the word models and \texttt{Yes-activation} in the nonword models did not reach significance for one of two training subjects, but their inclusion did substantially improve model fit. These predictors were therefore retained.  Further details on the validation modeling are provided in the Supplementary Materials.

\subsubsection{Regression with GAMs}

We used the Generalised Additive Model \citep[GAM;][]{hastie1987generalized,wood2011gam}, as implemented in the \textbf{mgcv} package for R, to study the functional relation between response latencies and our predictor variables. GAMs are regression models that can incorporate non-linear effects of one or more predictors on the response variable \citep[see also][]{baayen2017cave}.

The BLP dataset is too large to allow fitting with an insightful generalized additive mixed model. To avoid this computational bottleneck, we fitted separate GAMs to the data of the individual subjects. Furthermore, for ease of interpretation, we fitted separate models to the word data and to the nonword data.

The sequences of reaction times in the BLP form time series that are characterized by autocorrelations \citep[e.g.][]{baayen2017cave,baayen2022note}. GAMs can take autocorrelations into account by building an AR(1) process into the residuals, such that the residual at $t$ is a proportion $\rho$ of the residual at $t-1$ plus Gaussian noise. We obtained $\rho$ for each model individually by first extracting the autocorrelation values of residuals at lag 1 from a GAM without autocorrelation with classical predictors for both words and nonwords respectively. We then set this value as our $\rho$ for the subject, and ran both classical and DLM-based models, this time with the  autocorrelation parameter included. Note that the reaction times in our GAMs are not time series in the strictest sense, as we carried out separate analyses for words and nonwords as well as excluded extreme outliers (see above).

As the original BLP experiment was too long to perform all in one session, the participants were allowed to freely choose how many blocks they wanted to do in one day. A session expired after a break of more than 10 mins between blocks. Since we assumed that after such a break, a response would no longer be influenced by the previous one, we opted to restart the autocorrelation for each new session. We experimented with never restarting and restarting only for each new day of the experiment, but found that a session-based restart addressed the issue of inter-trial autocorrelation with greater precision for our validation data.

Model criticism revealed that the de-correlated residuals did not follow Gaussian distributions. As a consequence, our models remain approximate. To ensure that these approximate models are reasonable, we also considered Gaussian location-scale models, which model the effect of predictor variables on both mean and variance of the dependent variables, as well as Quantile GAMs, which are distribution free. The functional form of partial effects remained stable across these analyses. Full details are available in the Supplementary Materials.

We complemented the GAM analyses (Sections~\ref{sec:testing} and \ref{sec:learning}) with Linear Mixed Models (LMMs) fitted to the data of all subjects jointly, with one LMM fitted to the word data, and one to the nonword data.   Since  participant can be included as a random-effect factor, and by allowing interactions of participant with the other predictors, the LMM becomes  an eminent tool for studying individual differences between subjects.

Although it is in principle possible to use mixed GAMs,  for the large dataset of the BLP, we were confronted with two problems. First, the dataset is too big for the current implementation in \textit{mgcv} to estimate a model with the full complexity that we need. Second, a Generalised Additive Mixed Model with all necessary interactions, even if it were estimable, would be extremely difficult to interpret.  Therefore, to study individual variation within a regression framework, we needed to simplify.  The simplifying assumption that we made is that linear trends, although approximate, can be used to capture the main differences between participants.

The LMMs, which we fitted with the \textit{julia} package \textit{MixedModels.jl} \citep{bates2021mixed}, are reported in Section~\ref{sec:ind_diff}.

\section{Results and Discussion}\label{sec:results}

\noindent
In what follows, we first present our GAMs for both words and nonwords and show how well our predictors generalise across subjects. Based on these models we then address the main question of this study, namely whether trial-to-trial learning can be detected in the BLP data. Finally, we take a closer look at individual differences between subjects.

\subsection{Modeling reaction times to words and nonwords with GAMs}\label{sec:testing}

\subsubsection{Words}

\paragraph{GAM with Classical Predictors}
We started out by fitting a baseline model using only classical psycholinguistic measures (\texttt{Log Word frequency}, \texttt{Word length} and \texttt{Log Neighbourhood size}) to predict reaction times. This model cannot take trial-to-trial learning into account. Additionally, we included \texttt{Trial Number} and the participant's \texttt{Response} (word/nonword) as predictors. In the following, we will refer to the ratio of subjects for which an effect is significant ($\alpha=0.001$) as a predictor's ``reliability''\footnote{In addition, we also measured the contribution of each predictor to the overall AIC of a model for the first two validation subjects. The results can be found in the Supplementary Materials.}. An overview of the various predictors, the direction of their effect and reliability can be found in Table~\ref{tab:predictors_words_classical}.\footnote{Note that we found reliabilities to be similar in both of the disjunct word stimuli sets, and therefore report reliabilities for both sets jointly here. This applies to the GAM models with classical, non-incremental predictors as well as those based on incremental DLM-based predictors. Reliabilities split by stimuli list can be found in the Supplementary Materials.}

\begin{figure}
    \centering
    \includegraphics[width=\textwidth]{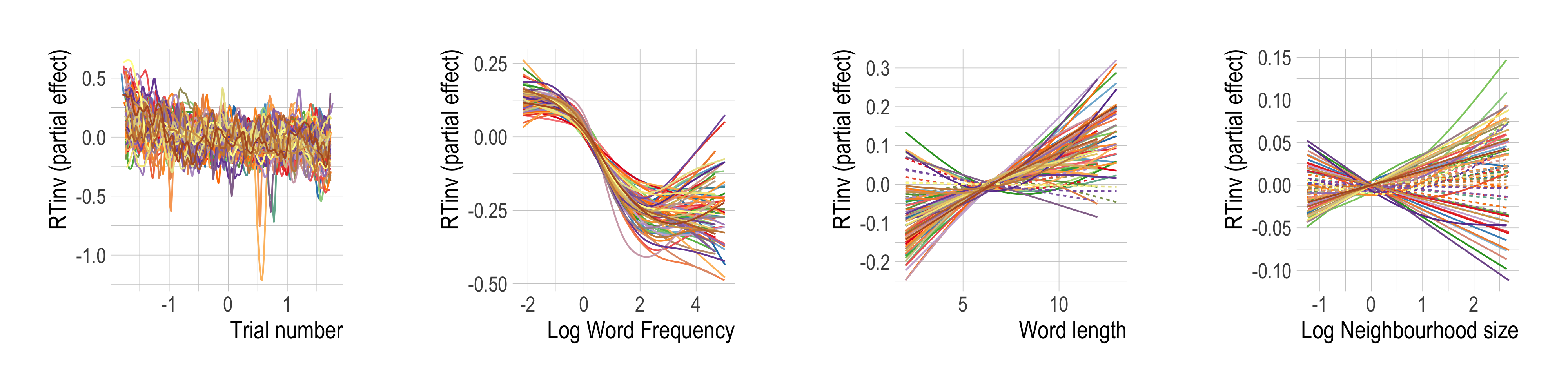}
    \caption{Partial effects of \textbf{classical predictors} for response latencies to \textbf{words} for all subjects. Solid lines are significant for  $\alpha = 0.001$. While the effects of \texttt{Log Word frequency} is very similar across all subjects, the effects of \texttt{Word length} and \texttt{Log Neighbourhood size} show substantial variability, indicative of widespread individual differences.}
    \label{fig:all_classical_words}
\end{figure}

\texttt{Trial number} was a significant predictor for all subjects. Inspecting the individual effects, we see that along the course of the experiment (see Figure~\ref{fig:all_classical_words}), reaction times generally became shorter, with a couple of exceptional subjects who remained relatively stable and others who even slowed down. There was also considerable variability within sessions \citep[cf.][]{baayen2017cave, pham2015vietnamese}. \texttt{Response} was significant for 82\% of participants.  \texttt{Log Word frequency} was also significant for all subjects. The effect was qualitatively remarkably similar across all subjects. Higher \texttt{Log Word frequency}  generally elicited shorter reaction times. At very high frequencies this effect was attenuated \citep{baayen2006morphological, keuleers2012british}. Higher \texttt{Word Length} (significant predictor for 87\% of subjects) gave rise to longer reaction times, except for five subjects for which the effect was U-shaped.  The U-shaped effect reported by \citet{baayen2005data} and \citet{new2006reexamining} apparently did not generalize to the majority of participants in the BLP. Finally, the most contested predictor for words, \texttt{Log Neighbourhood size}, was significant in 55\% of cases. The direction of the effect was incoherent across subjects. For 28 of the subjects, higher \texttt{Log Neighbourhood size} elicited longer reaction times, whereas for 14 subjects they were shorter. The effects for the remaining subjects either were not significant or had no clear direction (one subject). This variability is presumably one of the reasons why the effect of \texttt{Log Neighbourhood size} was found to be so inconsistent across previous studies \citep{andrews1992frequency, balota2004visual,keuleers2012british}. It should be noted that this does not invalidate the construct of neighbourhood size, but that it is important to better understand the reason for this variability.

\begin{table}[]
    \centering
    \begin{tabular}{l|l|r}
    \hline
        Predictor & Increase elicits \ldots &  Reliability\\
        \hline
        Trial number & shorter RTs (but wiggly) & 100\%\\
        Log Word frequency & shorter RTs, attenuated at high frequencies & 100\% \\
        Word length & longer RTs & 87\%\\
        Log Neighbourhood size & 65\% longer RTs, 33\% shorter RTs, rest U-shaped & 55\% \\
        \hline
        Response=W & 30\% shorter RTs, 70\% longer RTs & 82\%\\
        \hline
    \end{tabular}
    \caption{Predictors and their reliability for \textbf{words} in the \textbf{classical GAMs}. Effect of increase (given for significant predictors only) is intended as a summary and may differ for individual subjects (see Figure~\ref{fig:all_classical_words} for details). Reliability gives the percentage of subjects for which the predictor (regardless of direction) is significant ($p<0.001$). }
    \label{tab:predictors_words_classical}
\end{table}

\begin{figure}
    \centering
    \includegraphics[width=.75\textwidth]{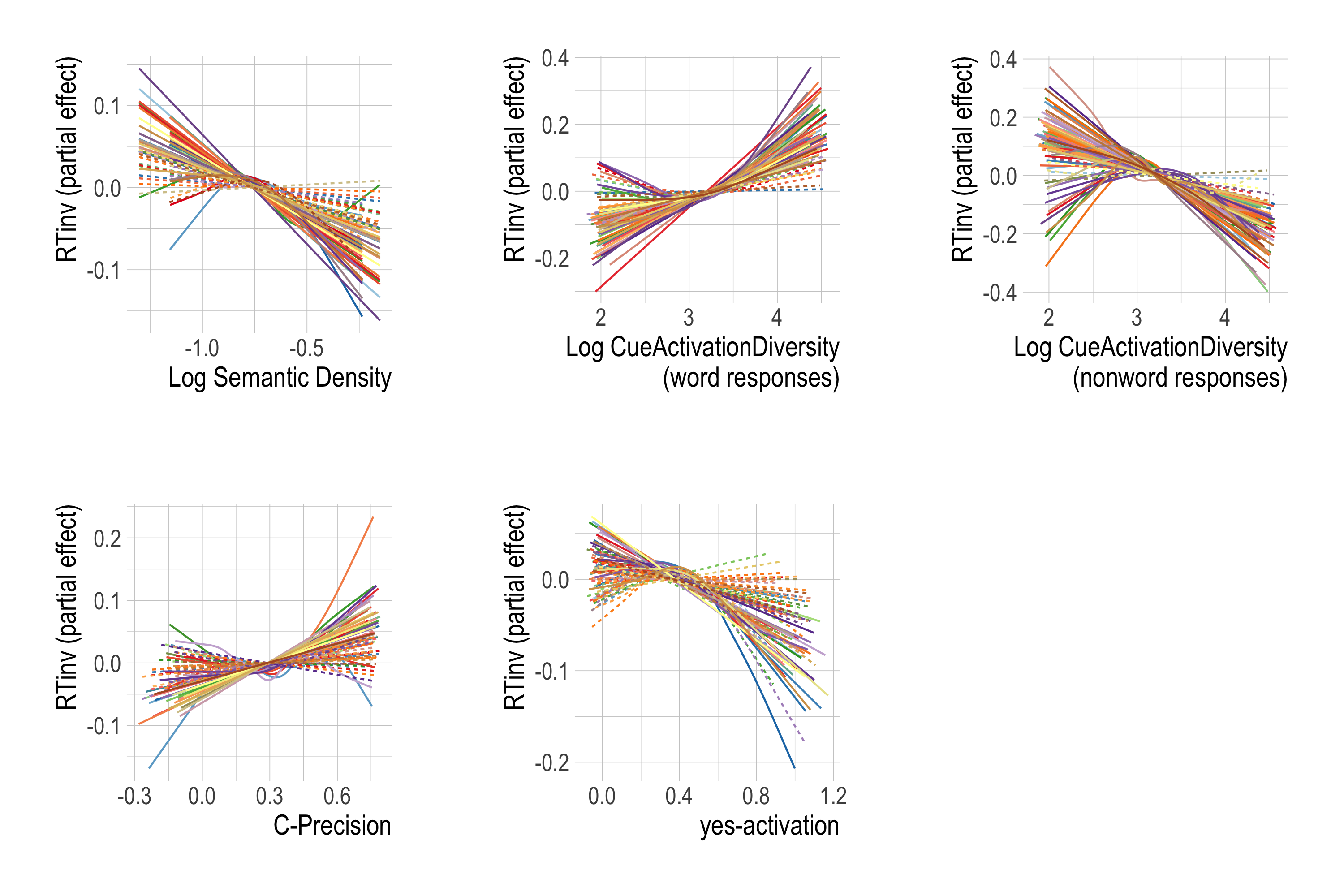}
    \caption{Partial effects of \textbf{DLM predictors} for all subjects (\textbf{words}). Solid lines have a significance level of $p<0.001$. Classical measures are omitted, as their partial effects are very similar to Figure~\ref{fig:all_classical_words}. The ranges of predictors vary within plots as a consequence of the between-subject design of the BLP.  Full figures can be found in the Supplementary Material.}
    \label{fig:all_ldl_words}
\end{figure}

\paragraph{GAM using DLM measures}
We included two sets of measures in our GAMs: a set of non-incremental measures (\texttt{Trial number}, \texttt{Log Word frequency}, \texttt{Word length} and \texttt{Response}), and four incremental measures from the DLM (\texttt{Log Semantic density}, \texttt{Log Cue Activation Diversity}, \texttt{C-Precision} and \texttt{Yes-activation}). The set of non-incremental measures does not include \texttt{Orthographic Neighbourhood Size} as by itself it has no clear theoretical motivation from the DLM perspective\footnote{But note that it is used implicitly in the \texttt{Form-driven Semantic Relatedness} measure for nonwords.}. Table~\ref{tab:predictors_words} provides an overview of the predictors and their reliability;  Figure~\ref{fig:all_ldl_words} visualises the partial effects.  The classical predictors in the dynamic GAMs had similar effects as in the baseline model, and are therefore not displayed (but see Supplementary Materials for further details).

\texttt{Trial Number} was significant across all subjects.  We included this predictor because effects which arise from e.g. increased motor training, task adaption or attention fluctuations \citep[cf.][]{baayen2022note} are outside the scope of our model. We note, however, that by including \texttt{Trial Number, we work against our hypothesis, as this predictor may absorb part of the effect of learning.}

As expected, \texttt{Log Word frequency} was again significant for all subjects. \texttt{Word length} was a significant predictor for somewhat fewer subjects (74\%), and \texttt{Response} for 77\%.

The partial effects of the predictors that are grounded in the DLM are visualized in Figure~\ref{fig:all_ldl_words}. \texttt{Log Semantic density} (top left) was significant for 56\% of all subjects: the denser the semantic space the predicted vector $\hat{\mathbf{s}}$ landed in, the faster the response. This fits well with insights gained with models such as MROM, where higher general activation implies higher lexicality --- and thus faster reaction times \citep{grainger1996orthographic}.

\texttt{Log Cue Activation Diversity} (top center), a measure of the uncertainty in the \chat \ vector,
had high reliability for both word responses (78\% of subjects) and nonword responses (90\% of subjects). If the response was nonword, higher \texttt{Log Cue Activation Diversity} was associated with shorter reaction times (i.e. high uncertainty led to faster reaction times for nonword responses), while for word responses it elicited longer reaction times (high uncertainty led to slower reaction times).

\texttt{C-Precision} (bottom left), which measures how correlated the predicted vector \chat \ is with the original form vector $\mathbf{c}$, was significant for about half of the subjects. For these subjects, the more precise the mapping back from the semantics to the form was, the longer reaction times were.
Our interpretation of this effect is that a well-supported form vector requires suppressing the production system more, which takes resources away from making a rapid lexicality decision.

The effect of  \texttt{Yes-activation} is displayed in the bottom center of Figure~\ref{fig:all_ldl_words}. Its effect was significant for 32\% of subjects. For these subjects, the more sublexical evidence in favour of a word outcome (higher \texttt{Yes-activation}) was available, the faster participants reacted.

\begin{table}[]
    \centering
    \begin{tabularx}{\textwidth}{X|X|r}
    \hline
        Predictor & Increase elicits \ldots &  Reliability\\
        \hline
        Trial number & shorter RTs (but wiggly) & 100\%\\
        Log Word frequency & shorter RTs, attenuated at high frequencies & 100\% \\
        Word length & longer RTs & 74\%\\
        Log Semantic density & shorter RTs & 56\% \\
        Log Cue Activation Diversity (word response) & longer RTs & 78\% \\
        Log Cue Activation Diversity (nonword response) & shorter RTs & 90\% \\
        C-Precision & longer RTs & 51\% \\
        Yes-activation & shorter RTs & 32\% \\
        \hline
        Response=W & 35\% shorter RTs, 65\% longer RTs & 77\%\\
        \hline
    \end{tabularx}
    \caption{Predictors and their reliability for \textbf{words} in the \textbf{DLM-based models}. Effect of increase (given for significant predictors only) is intended as a summary and may differ for individual subjects (see Figure~\ref{fig:all_ldl_words} for details). Reliability gives the percentage of subjects for which the predictor (regardless of direction) is  significant ($\alpha=0.001$).}
    \label{tab:predictors_words}
\end{table}

We finally observed that 99\% of the GAMs based on the DLM measures (with incremental updates) had a lower AIC value (i.e. better model fit) than the classical models (Mean AIC difference 152.6; see also Figure~\ref{fig:AIC_diff} below). In other words, the DLM-derived measures offer substantial additional precision to models based on the classical predictors only.

\subsubsection{Nonwords}

\begin{figure}
    \centering
    \includegraphics[width=.75\textwidth]{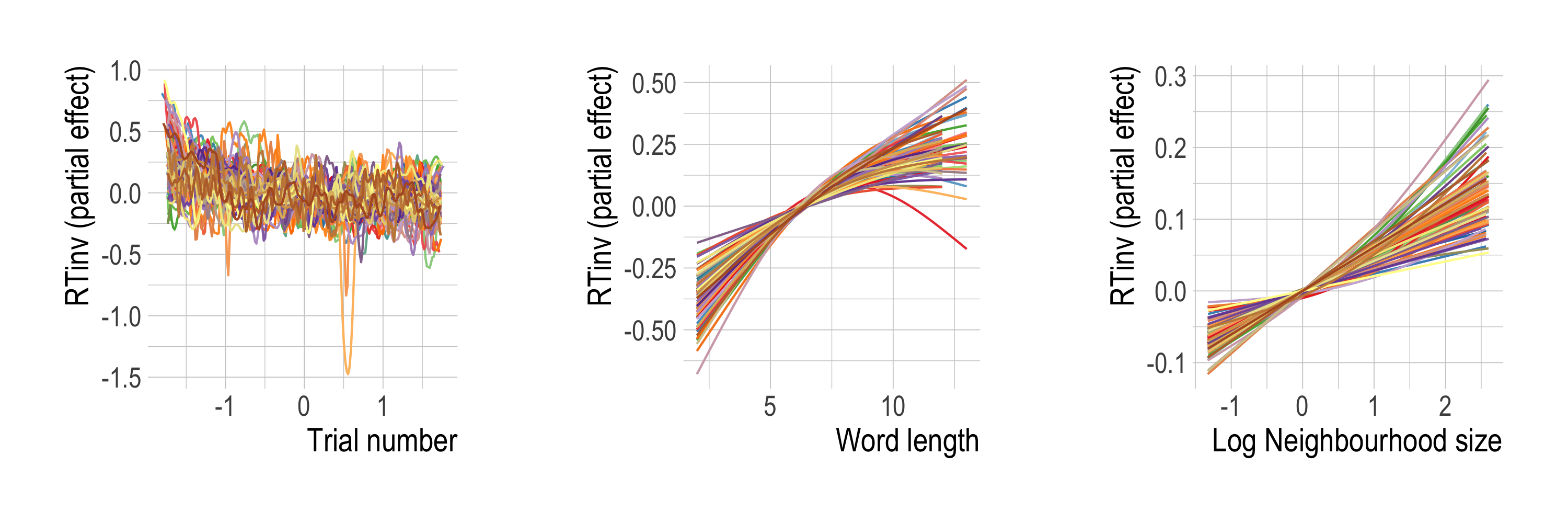}
    \caption{Partial effects of \textbf{classical predictors} in GAMs fitted to reaction times to \textbf{nonwords}, for all subjects. Solid lines are significant at  $\alpha = 0.001$. Effects are remarkably  uniform across subjects.}
    \label{fig:all_classical_nonwords}
\end{figure}

\paragraph{GAM with Classical predictors}
As we had no frequencies for the nonwords in the BLP \citep[but see][for the predictivity of nonword frequencies from the web for lexical decision latencies]{hendrix2021word}, we only included \texttt{Trial number}, \texttt{Word length}, \texttt{Log Neighbourhood size} and \texttt{Response} as classical predictors in our baseline model. Their effects are visualised in Figure~\ref{fig:all_classical_nonwords}.
Overall, reaction times tended to decrease for increasing \texttt{Trial Number}.  For both increasing \texttt{Word length}, and increasing \texttt{Log Neighbourhood size}, reaction times increased, replicating results from previous studies \citep{balota2004visual,yap2015responding}. All three covariates were significant for all subjects; the binary variable \texttt{Response} was significant for 82\% of subjects (Table~\ref{tab:predictors_nonwords_classical}).

\begin{table}[]
    \centering
    \begin{tabular}{l|l|r}
    \hline
        Predictor & Increase elicits \ldots &  Reliability\\
        \hline
        Trial number & shorter RTs (but wiggly) & 100\%\\
        Word length & longer RTs & 100\%\\
        Log Neighbourhood size & longer RTs & 100\% \\
        \hline
        Response=W & 16\% shorter RTs, 84\% longer RTs & 82\%\\
        \hline
    \end{tabular}
    \caption{Predictors and their reliability for reaction times to \textbf{nonwords} in the \textbf{classical GAMs}. Effect of increase (given for significant predictors only) is intended as a summary and may differ for individual subjects (see Figure~\ref{fig:all_classical_nonwords} for details). Reliability gives the percentage of subjects for which the predictor (regardless of direction) is significant ($\alpha = 0.001$).}
    \label{tab:predictors_nonwords_classical}
\end{table}

\begin{figure}[!ht]
    \centering
    \begin{subfigure}[b]{.5\textwidth}
    \centering
    \includegraphics[width=\textwidth]{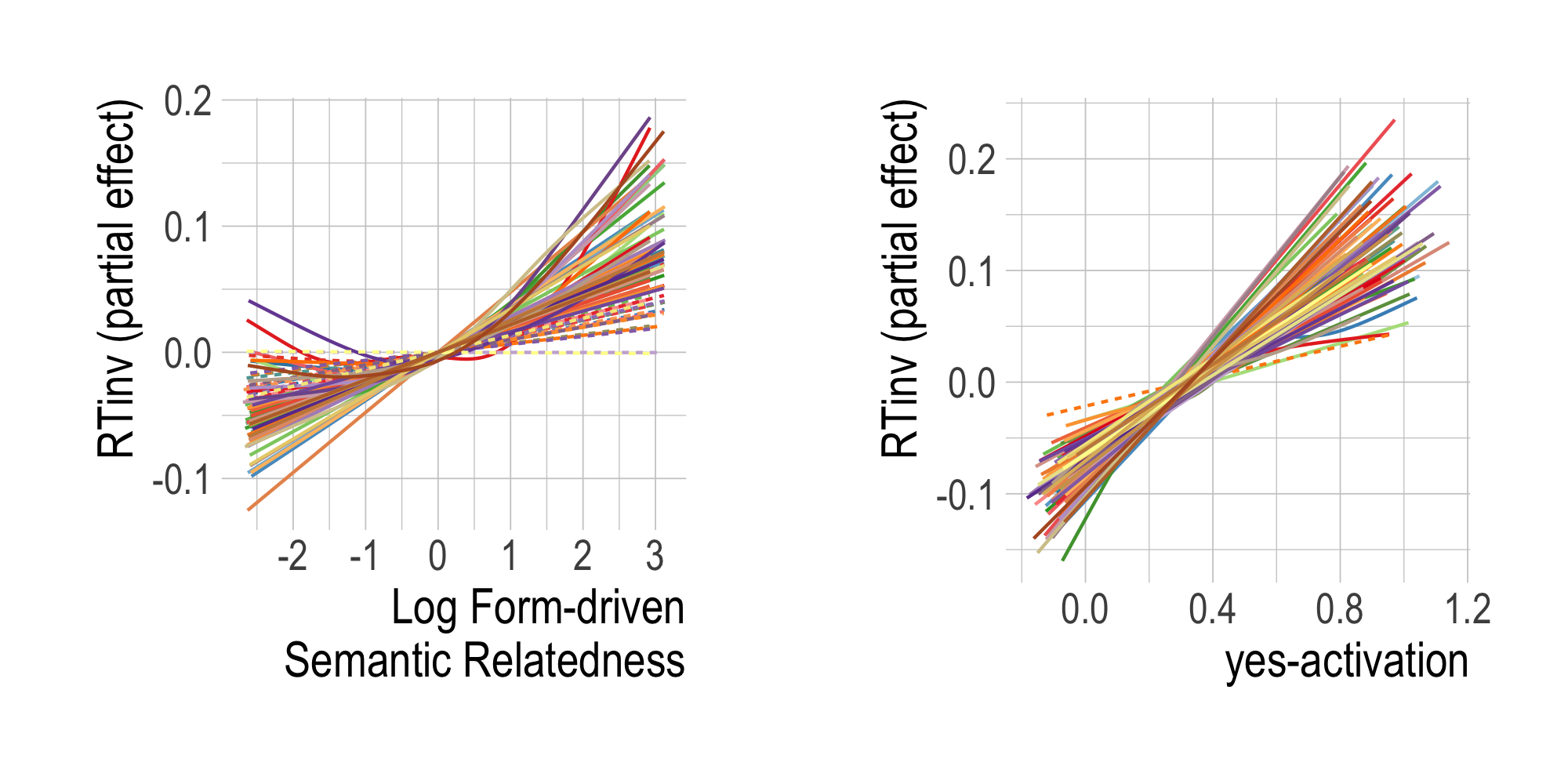}
    \caption{}
    \label{fig:all_ldl_nonwords_simple}
    \end{subfigure}
    \begin{subfigure}[b]{\textwidth}
    \centering
    \includegraphics[width=\textwidth, trim={0 35cm 0 0},clip]{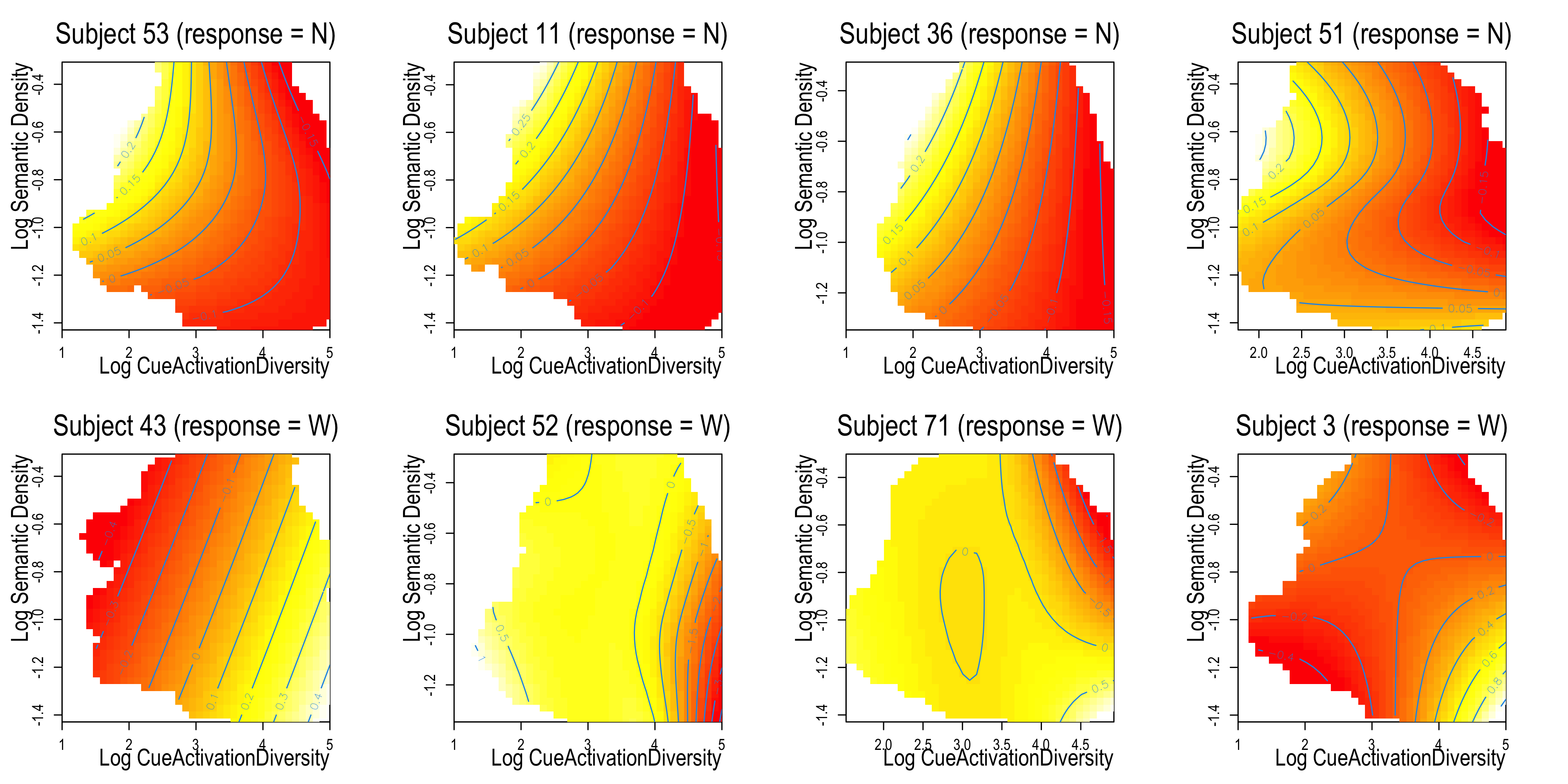}
    \caption{}
    \label{fig:all_ldl_nonwords_interaction}
    \end{subfigure}
    \caption{(a) Partial effects of the thin-plate regression smooth \textbf{DLM predictors} for all subjects (\textbf{nonwords}). Solid lines represent significance for $\alpha =  0.001$. Classical measures are omitted, as they are very similar to Figure~\ref{fig:all_classical_words}.
    (b) Sample of tensor product partial effect for nonwords with nonword responses, yellow indicates longer, and red shorter reaction times. Full figures, including those for word responses, can be found in the Supplementary Material.}
    \label{fig:all_ldl_nonwords}
\end{figure}

\paragraph{GAM including DLM measures}
Turning again to GAMs with DLM measures, we included \texttt{Trial number} and \texttt{Word length} as non-incremental predictors. We did not include \texttt{Neighborhood Size} as predictor as again, it does not have a theoretical motivation and is strongly correlated with \texttt{Form-driven Semantic Relatedness} (a measure that goes beyond a simple count of the number of close competitors).
Unsurprisingly, both \texttt{Trial number} and \texttt{Word length} were significant for nearly all subjects and had a remarkably similar effect as in the classical models (see Supplementary Materials for visualization). \texttt{Response} was significant for 88\% of subjects. Additionally, we found \texttt{Log Form-driven Semantic Relatedness}, \texttt{Yes-activation}, and an interaction between \texttt{Log Cue Activation Diversity} and \texttt{Log Semantic density} conditioned on response to be good predictors for nonword reaction times. These effects are summarised in Table~\ref{tab:predictors_nonwords} and visualised in Figure~\ref{fig:all_ldl_nonwords}.

The left panel in Figure~\ref{fig:all_ldl_nonwords_simple} shows the effect of \texttt{Log Form-driven Semantic Relatedness}, which was relatively reliable ($p<0.001$ for 71\% of subjects): The more orthographic neighbours of a nonword there were, and the further apart these neighbours were in semantic space (and hence the less confusable the meanings of these neighbors are), the longer it took a subject to react to the nonword. This finding dovetails well with the effect of \texttt{Log Neighbourhood size}, with which it is correlated ($r = 0.67$): the more orthographic neighbours a nonword has, the more it looks like a word, and the longer it takes to reject it as a word.

A higher \texttt{Yes-activation}, i.e. a higher support for a word outcome, predicted longer response latencies. As expected, its effect was opposite to its effect for words. While \texttt{Yes-activation} was only significant in 32\% of subjects for words, it was significant for virtually all subjects for nonwords (99\%).

One interaction emerged for the validation data, and turned out to be robust across all subjects, namely, an interaction between \texttt{Log Cue Activation Diversity} and \texttt{Log Semantic density} for nonword responses.   The left three panels in the upper row of Figure~\ref{fig:all_ldl_nonwords_interaction} present the regression surfaces (obtained with tensor product smooths) for Subjects 53, 11 and 36. These subjects show the pattern that was typical for most subjects: higher \texttt{Log Cue Activation Diversity} elicited shorter reaction times, while higher \texttt{Log Semantic density} elicited longer reaction times specifically for lower values of \texttt{Log Cue Activation Diversity}.  Subject 51 (upper right panel) shows a somewhat wiggly effect of \texttt{Log Semantic density} that is less characteristic of the full set of subjects. Plots for all subjects can be found in the Supplementary Materials.

We note here that for word responses, an interaction between \texttt{Log Cue Activation Diversity} and \texttt{Log Semantic density} was only significant for 22\% of subjects, and was highly variable and inconsistent across subjects.

\begin{table}[!ht]
    \centering
    \begin{tabularx}{\textwidth}{X|X|r}
    \hline
        Predictor & Increase elicits... &  Reliability\\
        \hline
        Trial number & shorter RTs (but wiggly) & 100\%\\
        Word length & longer RTs & 99\%\\
        Form-driven Semantic Relatedness & longer RTs & 71\% \\
        Yes-activation & longer RTs & 99\% \\
        N response: Sem. Density x Cue Activation Diversity & Cue Activation Diversity shorter RTs,  Sem. Density longer RTs (effect stronger for lower Cue Activation Diversity)  & 100\% \\
        W response: Sem. Density x Cue Activation Diversity & no generalisable effect in any direction & 22\%\\
        \hline
        Response=W & 13\% shorter RTs, 87\% longer RTs & 88\% \\
        \hline
    \end{tabularx}
    \caption{Predictors and their reliability for \textbf{nonwords} in the \textbf{DLM-based GAM models}. Effect of increase (given for significant predictors only) is intended as a summary and may differ for individual subjects (see Figure~\ref{fig:all_ldl_nonwords} for details). Reliability gives the percentage of subjects for which the predictor (regardless of direction) is significant ($\alpha = 0.001$).}
    \label{tab:predictors_nonwords}
\end{table}

Finally, the GAMs for nonwords based on DLM measures had a lower AIC (i.e. better model fit) than the classical models for all subjects (Mean  AIC difference 135.3; see also Figure~\ref{fig:AIC_diff}). DLM measures seem therefore well suited to also predict nonword reaction times.

\subsection{Trial-to-trial learning}\label{sec:learning}

In order to answer the main question of this study, whether the modelling profits from incremental updates during the simulation, we ran
an additional model for each subject, using the DLM-based predictors, but without ever updating these from trial to trial.
%%%%%%%%%%%%%The model specifications for the GAMs were identical, with one exception: the \texttt{Yes-Activation} measure is not available for the model without trial-by-trial learning (henceforth the static models), and was therefore used as a predictor only for the models with  predictors that evolve with learning (henceforth the dynamic models).
This allowed us to directly compare, for any given subject, the contributions of measures obtained from a model with and a model without trial by trial learning.

\begin{figure}[!ht]
    \centering
    \begin{subfigure}[b]{\textwidth}
    \centering
    \includegraphics[width=\textwidth]{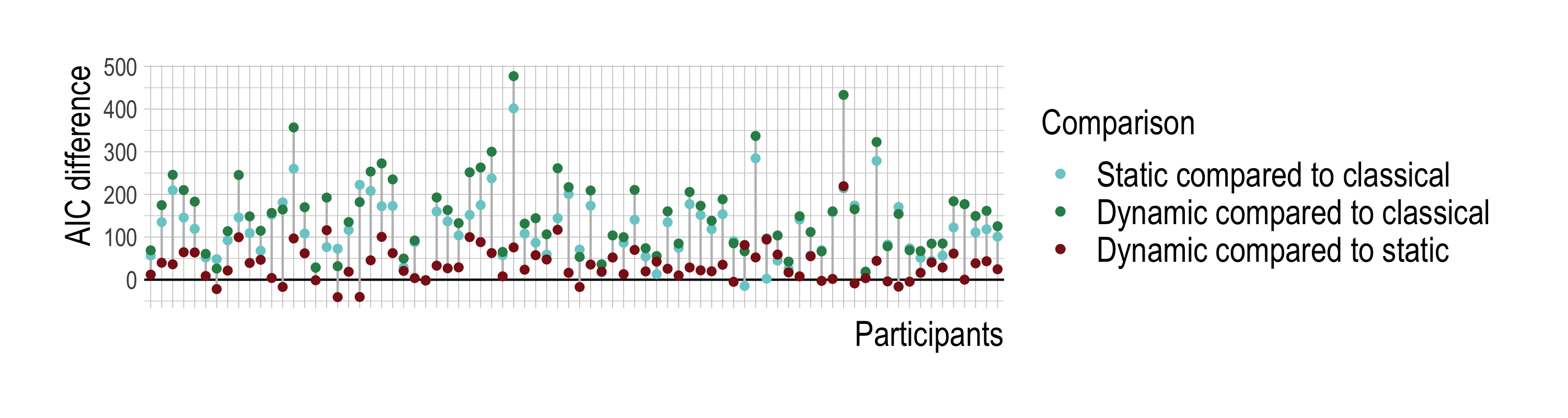}
    \caption{Words}
    \label{fig:AIC_diff_words}
    \end{subfigure}
    \begin{subfigure}[b]{\textwidth}
    \centering
    \includegraphics[width=\textwidth]{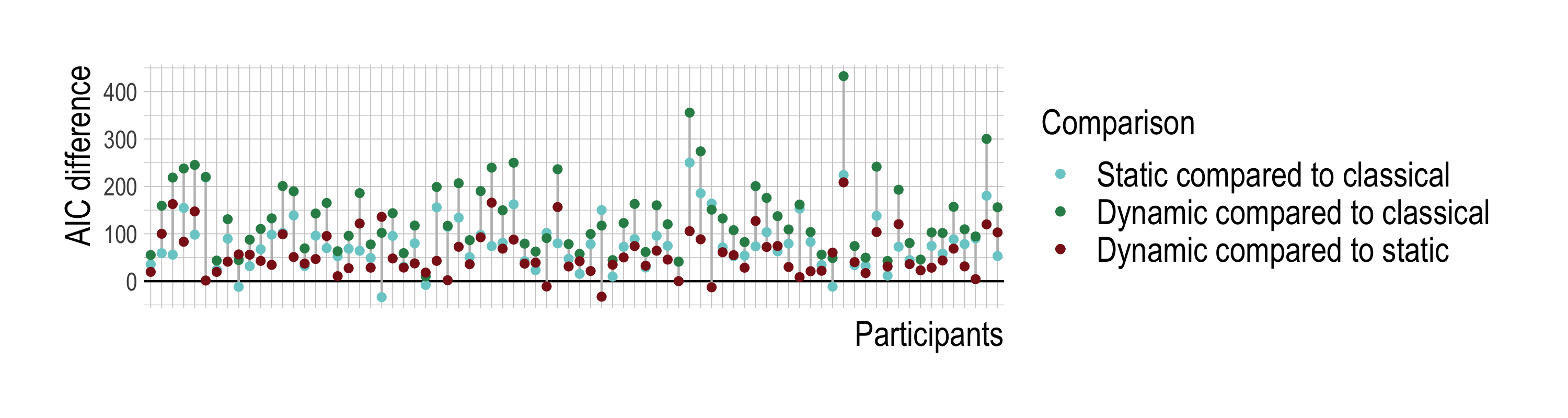}
    \caption{Nonwords}
    \label{fig:AIC_diff_nonwords}
    \end{subfigure}
    \caption{AIC comparisons of classical, static (i.e. no trial-to-trial learning) and dynamic (with trial-to-trial learning) models for both words (a) and nonwords (b). If, for example, the AIC difference of ``static compared to classical'' (turquoise) is positive for a subject, the static GAM has a better model fit than the classical one for this particular subject. The other comparisons can be interpreted analogously. Static and dynamic models almost always have higher relative likelihood than the classical model. Dynamic models mostly show a better model fit than static models, implying that the models benefit from trial-to-trial learning.}
    \label{fig:AIC_diff}
\end{figure}

The dynamic models for words had lower AIC values than the corresponding static models in 85\% of cases. Differences in AIC values ranged from -40.9 (static better than dynamic) to 219.2 (dynamic better than static) (\textit{M} 35.2). On average, the relative likelihood of dynamic compared to static models was $5.0 \times 10^{45}$. For nonwords, dynamic models were better than static ones in 94\% of cases (differences in AIC: \textit{M} 55.7, range -32.7 to 208.5), with the relative likelihood of dynamic compared to the static models on average $2.5 \times 10^{43}$. The differences in AIC values are presented in Figure~\ref{fig:AIC_diff}. A possible explanation for some of the simulations not profiting from trial-to-trial learning is that some of these respective subjects did learn trial by trial, but the learning rate we chose was so suboptimal, that their behaviour was better approximated by the measures based on the static models, rather than the learning ones.

Recall that for static models we cannot include the \texttt{Yes-Activation} predictor, as it is critically dependent on incremental updates. This raises the question of whether the improved model fit of dynamic simulations was due to the incremental updates of the main mapping matrices $\mathbf{F}$ and $\mathbf{G}$ during the simulation, or whether it was mainly the \texttt{Yes-Activation} that was responsible for improving goodness of fit. To investigate this possibility, we ran GAMs for the dynamic simulations without \texttt{Yes-activation}
and again compared AIC values. We found that for word models, even without \texttt{Yes-Activation}, dynamic GAMs still provided a better model fit for 82\% of the subjects, a reduction of a mere 3\% (\textit{M} AIC difference: 35.2). For nonwords, however, this was only the case for 60\% of the subjects, a reduction by 34\% (\textit{M} AIC difference: 3.0). Apparently, for responses to words, trial-to-trial updating of the lexical networks contributed substantially to the goodness of fit.  However, for nonwords, improvements in goodness of fit are to a much larger extent due to purely form-based sublexical learning.

\subsection{Individual differences}\label{sec:ind_diff}

In order to clarify the main differences between individual participants, we fitted an LMM to the reaction times for words, and a second LMM to the reaction times for nonwords.  These models included by-participant random intercepts as well as by-participant random slopes for all predictors. The random effect components of the two LMMs are summarized in  Table~\ref{tab:LMMs}.
The LMMs confirmed the direction of effect for all predictors, which all were well-supported ($p<0.001$), including predictors such as \texttt{Yes-activation} with relatively low reliability in the individual GAMs. Exceptions were the main effects as well as the interaction of \texttt{Log Semantic density} and \texttt{Log Cue Activation Diversity} for word responses in the nonword model ($p>0.68$). Possibly, this was because only 5.7\% (63,274) of responses to nonwords were word responses. Additionally, the table includes information on how important the individual random slope adjustments are to the overall model fit, by providing the AIC difference that removing the respective random slope adjustment would result in.

\begin{figure}
    \centering
    \begin{subfigure}[b]{\textwidth}
    \centering
    \includegraphics[width=\textwidth]{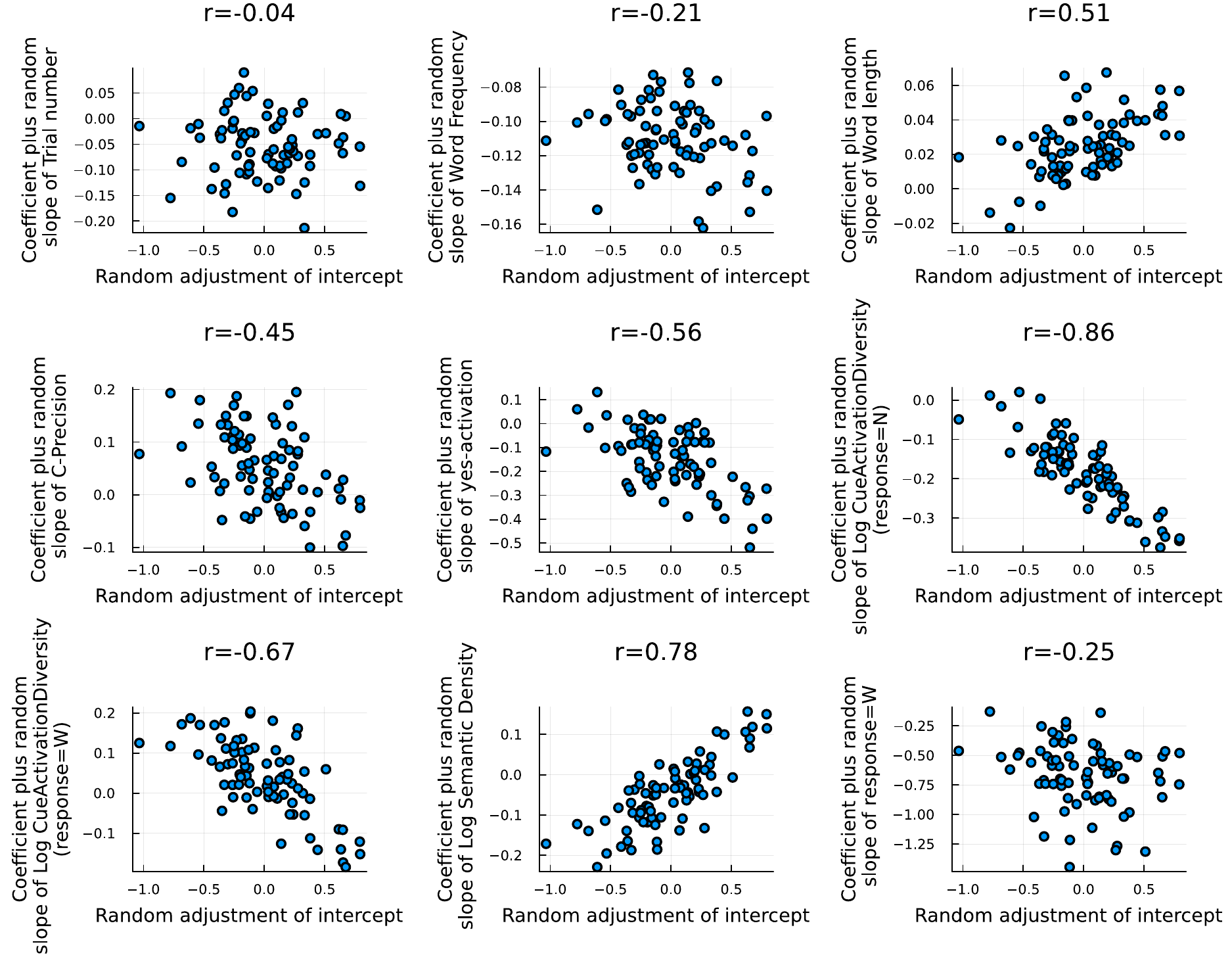}
    \caption{Words}
    \label{fig:re_plots_words}
    \end{subfigure}
    \begin{subfigure}[b]{\textwidth}
    \centering
    \includegraphics[width=\textwidth]{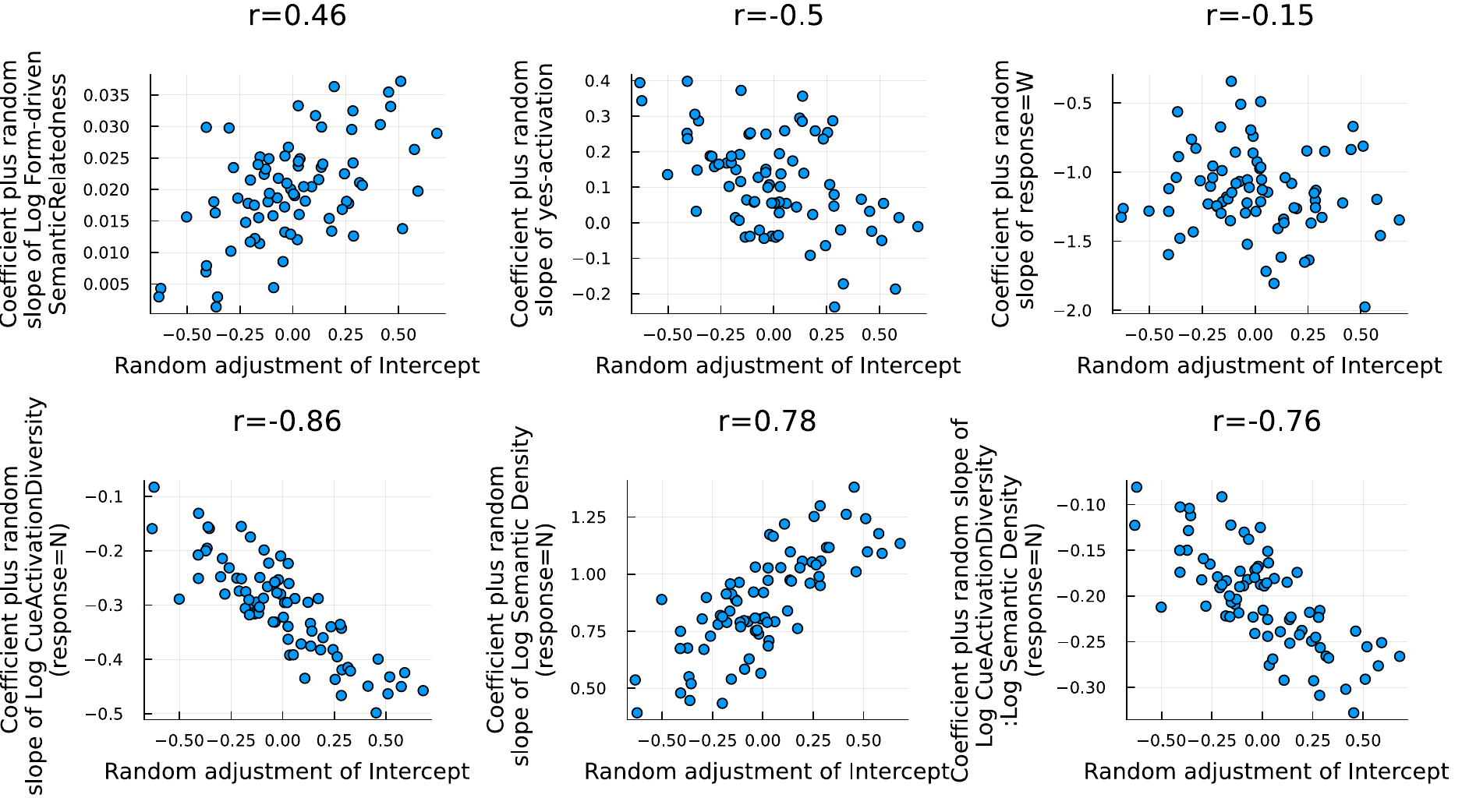}
    \caption{Nonwords}
    \label{fig:re_plots_nonwords}
    \end{subfigure}
    \caption{Correlation between selection of participant-specific coefficients and random adjustment of intercept.}
    \label{fig:re_plots}
\end{figure}

To understand subject-specific differences in the effects of our predictors, we make use of visualization by plotting by-subject random slopes against by-subject random intercepts (Figure \ref{fig:re_plots}).  The random intercepts represent the deviation of the average response time of the individual participants from the population mean response time, with slower subjects more to the right, and faster subjects more to the left in the scatterplots in Figure \ref{fig:re_plots}.

First consider individual variability as revealed for the three non-incremental, classical predictors by the scatterplots in the top row of  Figure~\ref{fig:re_plots_words} (full correlation tables can be found in the Supplementary Materials).  In this figure, the y-axis concerns the participant-specific coefficients of a given predictor, i.e., the population slope $+$ the participant-specific random slope (posterior mode).

For \texttt{Trial number} (upper left), there is no clear correlation between random intercept and slope adjustment. For the vast majority of participants, the slope of \texttt{Trial number} is negative: as the experiment proceeds, participants respond more quickly.
\texttt{Log Word frequency} (upper center) shows a weak correlation (-.21) for slopes and random intercepts, suggesting that possibly slower subjects have a stronger effect of frequency \citep[more negative slopes; see also][]{kuperman2013reassessing}.
For \texttt{Word length} on the other hand, a clear correlation is present.  For fast subjects (left side of the plot),
the effect of \texttt{Word length} is weak or even negative, whereas for slow subjects (right side of the plot), greater word length clearly predicts longer reaction times.  The correlations within the nonword model for \texttt{Trial number} and \texttt{Word Length} are very similar and not displayed in Figure~\ref{fig:re_plots_nonwords}, but further information is available in the Supplementary Materials.   In summary, of the classical predictors, a strong correlation with response speed is present only for \texttt{Word Length}.

Scatterplots for the DLM-based predictors are shown in the second and third row of Figure~\ref{fig:re_plots_words} for words, and in Figure~\ref{fig:re_plots_nonwords} for nonwords.
For words (Figure~\ref{fig:re_plots_words}), faster subjects (with a random intercept $< 0$) show large positive slopes for \texttt{C-Precision} (center left panel). For slower participants, the slope is close to zero, and sometimes negative.  A stronger correlation is visible for \texttt{Yes-Activation} (center middle panel). For almost all subjects, the slope is negative, and more so for slower participants. An even stronger correlation emerges for \texttt{Log Cue Activation Diversity} when participants are executing nonword responses to word stimuli (center right panel).  There is hardly an effect for the fastest subjects, but the slower a participant responds on average, the more negative the slope of \texttt{Log Cue Activation Diversity} becomes.  When executing a word response, a clear negative correlation is also present (bottom left panel), but the datapoints are shifted upwards, with large positive slopes for the fastest participants, and only the very slowest subjects showing negative slopes.  A strong positive correlation characterizes participants response behavior with respect to \texttt{Log Semantic Density}. The fastest participants show facilitation, but this effect reverses into inhibition for the slowest responders.  Finally, there is no clear correlation of the slope of \texttt{Response} and random intercept (lower right panel).

The left-hand part of Table~\ref{tab:ind_diff_table} contrasts the effects for the slower subjects as opposed to the faster subjects.  The pattern that emerges is that slow subjects are primarily making use of words' form properties. They take more time to respond to longer words, and they speed up when the orthographic features of the word provide good support for a yes response. When making a nonword response, they decide more quickly when the uncertainty of the predicted form vector is greater (\texttt{Log Cue Activation Diversity}).  By contrast, the faster participants respond faster for words with greater semantic density, and they do not show much of an effect of word length. Faster responders appear to focus more on meaning. This explains why they respond more slowly when  \texttt{C-Precision} is high:  When the semantics precisely maps back onto form, faster responders are distracted by supporting evidence from words' forms, having to suppress saying the word out loud.  When making word responses, they also respond more slowly when \texttt{Log Cue Activation Diversity} is high, indicating that uncertainty about the form space is also detrimental for faster participants when presented with words.

Further insight into the individual differences between faster and slower responders is provided by the correlations in the random effects for nonwords, which are presented in Figure~\ref{fig:re_plots_nonwords}. A greater \texttt{Log Form-driven Semantic Relatedness} invariably led to longer nonword decisions (upper left panel), especially for the slower subjects. For the fastest subjects, there hardly was any effect.

Just as for responses to words, a negative correlation was present for the slopes of \texttt{Yes-Activation} and the random intercepts (upper center panel).  But whereas for words, all but the fastest responders had negative slopes, for nonwords, negative slopes were present only for the slowest responders.  In other words, slow responders made use of yes-activation to speed up their responses to words and hardly made use of yes-activation for nonwords; however, fast responders did not use yes-activation much for words, and were slowed down by this information for nonwords.

When participants made a nonword response to nonword stimuli, the slope of \texttt{Log Cue Activ\-at\-ion Diversity} was always negative, and more so for slower responders (lower left panel).  The same negative correlation was present for word responses to words. But whereas for words, responses were slowed down most for faster  subjects, for nonwords, responses were speeded up more for slower responders.  Apparently, uncertainty about the form predicted by the feedback loop differentially affected participants' response strategies for words and nonwords.

When making nonword responses, stimuli with a greater \texttt{Semantic Density} (center bottom panel) elicited longer reaction times, especially for the slower responders.  For nonwords, even the fast responders have positive slopes for semantic density. By contrast, for words, fast responders had negative slopes.  Apparently, fast responders used dense semantic neighborhoods to respond more quickly to words, at a small cost for response speed for nonwords. Conversely, the slowest participants were hardly slowed down for words, but were especially slow to respond to nonwords in dense semantic neighborhoods.

The effects of \texttt{Log Cue Activation Diversity} and \texttt{Log Semantic Density} were modulated by a multiplicative interaction (bottom right panel). The slowing effect of \texttt{Log Semantic Density} is attenuated by higher values of \texttt{Log Cue Activation Diversity},
and most prominently so for the slower responders. Correspondingly, the negative effect of \texttt{Log Cue Activation Diversity} on reaction times is enhanced by higher values of \texttt{Log Semantic Density}, again especially so for slower responders.
(For the joint effect of these predictors according to the GAM, see the example contour plots for subjects 11 and 36 in Figure~\ref{fig:all_ldl_nonwords}; the estimated surfaces for these subjects are similar to sections of a hyperbolic plane, the surface modeled with a multiplicative interaction in the LMM.)

\begin{table}[]
    \centering
    \begin{subfigure}[b]{\textwidth}
    \centering
    \begin{tabularx}{\textwidth}
{X | r | r | r | r | r | r}
 & Est. & SE & z & p & $\sigma_\text{participant}$ & AIC diff $\sigma$\\
\hline
(Intercept) & -1.1370 & 0.0441 & -25.81 & $<0.0001$ & 0.3775 & \\\cline{7-7}
Trial number & -0.0531 & 0.0067 & -7.98 & $<0.0001$ & 0.0586 & 14,931\\\cline{7-7}
in\_bnc=1 & -0.1248 & 0.0060 & -20.92 & $<0.0001$ & 0.0357 & 37\\\cline{7-7}
Word length & 0.0246 & 0.0020 & 12.23 & $<0.0001$ & 0.0176 & 3,118\\\cline{7-7}
C-Precision & 0.0544 & 0.0094 & 5.81 & $<0.0001$ & 0.0764 & 353\\\cline{7-7}
Yes-activation & -0.1517 & 0.0151 & -10.05 & $<0.0001$ & 0.1302 & 1,472 \\\cline{7-7}
Log Semantic density & -0.0450 & 0.0103 & -4.35 & $<0.0001$ & 0.0870 & 703\\\cline{7-7}
response=W & -0.6712 & 0.0330 & -20.36 & $<0.0001$ & 0.2817 & 936\\\cline{7-7}
Log Word frequency \& in\_bnc & -0.1100 & 0.0023 & -47.74 & $<0.0001$ & 0.0200 & \multirow{2}{*}{1,869}\\\cline{7-7}
Log Cue Activation Diversity (response=N) & -0.1832 & 0.0108 & -16.90 & $<0.0001$ & 0.0931 & \multirow{4}{*}{2,817}\\
Log Cue Activation Diversity (response=W) & 0.0380 & 0.0107 & 3.55 & 0.0004 & 0.0928 &\\\cline{7-7}
Residual & 0.3883 &  &  &  &  &\\
\end{tabularx}
\caption{Words}
    \end{subfigure}\hfill
\begin{subfigure}[b]{\textwidth}
\centering
\begin{tabularx}{\textwidth}
{X | r | r | r | r | r | r}
 & Est. & SE & z & p & $\sigma_\text{participant}$ & AIC diff $\sigma$\\
\hline
(Intercept) & -0.8557 & 0.0423 & -20.21 & $<0.0001$ & 0.3237 &\\\cline{7-7}
Trial number & -0.0628 & 0.0069 & -9.08 & $<0.0001$ & 0.0610 & 20,621\\\cline{7-7}
Word length & 0.0470 & 0.0018 & 26.37 & $<0.0001$ & 0.0154 & 1,686\\\cline{7-7}
has\_neighbours\_path=1 & 0.0718 & 0.0026 & 27.12 & $<0.0001$ & 0.0218 & 406\\\cline{7-7}
Yes-activation & 0.1135 & 0.0162 & 6.99 & $<0.0001$ & 0.1406 & 1,726\\\cline{7-7}
response=W & -1.1454 & 0.0983 & -11.65 & $<0.0001$ & 0.3793 & 9\\\cline{7-7}
Log Form-driven Semantic Relatedness \& has\_neighbours\_path & 0.0198 & 0.0012 & 16.90 & $<0.0001$ & 0.0091 & \multirow{3}{*}{140}\\\cline{7-7}
Log Cue Activation Diversity (response=N) & -0.3071 & 0.0129 & -23.75 & $<0.0001$ & 0.1016 & \multirow{13}{*}{1,821}\\
Log Cue Activation Diversity (response=W) & 0.0020 & 0.0294 & 0.07 & 0.9450 & 0.1306 &\\
Log Semantic density (response=N) & 0.8793 & 0.0366 & 24.05 & $<0.0001$ & 0.2541 &\\
Log Semantic density (response=W) & 0.0119 & 0.1090 & 0.11 & 0.9134 & 0.4063 &\\
Log Cue Activation Diversity \& Log Semantic density (response=N) & -0.2038 & 0.0096 & -21.30 & $<0.0001$ & 0.0630 &\\
Log Cue Activation Diversity \& Log Semantic density (response=W) & -0.0133 & 0.0322 & -0.41 & 0.6795 & 0.1212 &\\\cline{7-7}
Residual & 0.3841 &  &  &  &  &\\
\end{tabularx}
\caption{Nonwords}
\end{subfigure}
    \caption{Intercepts and random slopes estimated by the LMMs for words (a) and nonwords (b). Factors are treatment-coded. ``AIC diff $\sigma$'' indicates the change in AIC if the pertinent random slope is removed.}
    \label{tab:LMMs}
\end{table}

\begin{table}[!ht]
    \centering
    \begin{tabularx}{\textwidth}{c||Z|Z||Z|Z}
        \hline
         & \multicolumn{2}{l||}{\textbf{Words}} & \multicolumn{2}{l}{\textbf{Nonwords}} \\
         \hline
         & strong effect & weak/reversed effect & strong effect & weak/reversed effect \\
         \hline
        \textbf{Slower subjects} & \texttt{Word length}, \texttt{Yes-activation}, \texttt{Log Cue Activation Diversity (response=N)} &\texttt{Log Cue Activation Diversity (response=W)}, \texttt{Log Semantic density}, \texttt{C-Precision} &\texttt{Word length}, \texttt{Log Semantic density (response=N)}, \texttt{Log Form-driven Semantic Relatedness}, \texttt{Log Cue Activation Diversity (response=N)}, \texttt{Log Cue Activation Diversity:Log Semantic density (response=N)} & \texttt{Yes-activation}\\
        \hline
        \textbf{Faster subjects} & \texttt{Log Cue Activation Diversity (response=W)}, \texttt{Log Semantic density}, \texttt{C-Precision}  & \texttt{Word length}, \texttt{Yes-activation}, \texttt{Log Cue Activation Diversity (response=N)} & \texttt{Yes-activation} & \texttt{Word length}, \texttt{Log Semantic density (response=N)},  \texttt{Log Cue Activation Diversity} (response=N), \texttt{Log Form-driven Semantic Relatedness}, \texttt{Log Cue Activation Diversity:Log Semantic density (response=N)}\\
        \hline
    \end{tabularx}
    \caption{Summary table of individual differences in predictor strengths, for slower and faster subjects.
    Faster subjects optimize their responses by focusing on meaning for words (leading to delays from form measures), and on form for nonwords.  Slower subjects optimize their responses by focusing on form across words and nonwords. In addition, for nonwords, slower subjects suffer from interference from semantics.
    }
    \label{tab:ind_diff_table}
\end{table}

The right part of Table~\ref{tab:ind_diff_table} summarizes the effects for nonwords, contrasting faster and slower responders. The pattern that emerges from Table~\ref{tab:ind_diff_table} is the following.  Slower subjects focus primarily on form properties.  Slower subjects are especially slow for longer words and longer nonwords.  For words, they respond extra fast for greater \texttt{Yes-activation}, another measure of form. For nonwords, a greater uncertainty about the form predicted by the feedback loop allows slower subjects to respond especially quickly. At the same time, measures of meaning predict especially elongated response times for slower responders (\texttt{Log Form-driven Semantic Relatedness, Log Semantic density}). This suggests that the slower subjects are attempting to make sense of nonwords' semantics, but that this slows them down.
Faster subjects, on the other hand, do not reveal solid effects of word length. When faster subjects are responding to words, they show more facilitation for \texttt{Log Semantic Density}, and more inhibition for two form measures, \texttt{C-Precision} and \texttt{Log Cue Activation Diversity}.  For nonwords, faster subjects have larger positive slopes for \texttt{Yes-Activation}: when a stimulus' form features provide better support for a word decision, faster subjects are slowed down more in their responses.

In other words, faster subjects optimize their responses by focusing on meaning for words (leading to delays from form measures), and on form for nonwords.  Slower subjects optimize their responses by focusing on form across words and nonwords. In addition, for nonwords, slower subjects suffer from interference from semantics.

In the end, the question remains whether we can link these individual differences to hyperparameters of the DLM. Theoretically, there are three such parameters which were held constant across all participants: first, the prior knowledge of the model before entering the simulation was the same for all subjects, but it is likely that differing prior knowledge gives rise to some individual differences \citep[e.g.][]{kuperman2013reassessing}. Secondly, the two learning rates for \texttt{Yes-Activation} and for ``lexical'' learning were held constant for all participants. It is possible that for some subjects, the effects of our measures could not play out because they were based on suboptimal learning rates. While we have not explored these hyperparameters in-depth in the present study, the presented effects can inform future work investigating subject-specific hyperparameters underlying individual differences.

%\clearpage

\section{Conclusion}\label{sec:discussion}

We set out with the hypothesis that humans' lexical knowledge is continuously changing according to our experiences and environment. More specifically, we proposed that humans continuously learn and update their mental lexicons, from word use to word use. Since it is not clear how this effect can be measured in daily language use, we focused on detecting the effects of this continuous learning in psycholinguistic experiments. Therefore, the
main question of this study was whether effects of within-experiment learning are present in the lexical decision task, and can be detected using incremental learning. We investigated this question by predicting the lexical decision latencies of individual participants in the British Lexicon Project \citep[BLP;][]{keuleers2012british}, using the Discriminative Lexicon Model \citep[DLM;][]{baayen2019discriminative} to simulate trial-to-trial learning.  We then used predictors from this model to predict reaction times using Generalised Additive Models (GAMs). We found that our DLM-based measures provided a better model fit than ``traditional'' predictors (i.e. word frequency, length and orthographic neighbourhood density) to the reaction time data, from which we concluded that measures based on the DLM are indeed able to account for variance in lexical decision reaction times. We then hypothesised that measures based on DLM simulations with trial-to-trial learning updates would provide a better model fit than measures from non-learning DLM simulations. We found that for the majority of subjects, these learning-based predictors account for substantially more variance in reaction times than the corresponding predictors derived from a static DLM without incremental learning. We therefore conclude that trial-to-trial learning effects on reaction times are indeed present in the BLP, and that they can be detected with error-driven learning.

Our findings have several implications for theories of lexical representation and processing. First, several studies \citep[e.g.,][]{baayen2022note,balota2018dynamic, lima1997sequential, perea2003sequential} have documented inter-trial effects and discussed the consequences of these effects for the interpretation of experimental results. One particularly salient example is the study by \citet{palmeri2015experimental} on perceptual categorisation. In experiments where participants are supposed to learn categories by repeatedly categorising various stimuli, supposedly storing category representations in their long-term memory, participants can avoid learning these categories by using a ``relative judgment strategy'': Participants can simply base their judgments on the difference between the current and the previous stimulus in the task. The resulting behaviour is indistinguishable from behaviour where participants store learned categories in long-term memory. Basing conclusions about long-term memory on the results of such an experiment poses the danger of building a theory about long term learning on inter-trial effects.  The present study shows that there is a category learning component to lexical decision making as well. In the course of the experiment, participants learn to predict word/nonword status from words' sublexical features, bypassing long-term lexical knowledge in lexical memory altogether, and use this information to inform their lexicality decision alongside information from long-term lexical knowledge. This is mediated by individual differences: for example, faster-responding participants emerged as primarily making nonword decisions based on this task-specific predictor, while relying primarily on long-term lexical knowledge for word decisions.  The modeling framework that we are proposing makes it possible to tease apart these task effects from truly lexical effects.

Second, our results support the possibility that   our lexical knowledge is not static, but changes continuously as part of the never-ending adaptation of our cognitive systems to the environment \citep{hoffman2019case}. This has two implications: first, it supports learning-based theories of lexical processing \citep[e.g.][]{harm2004computing, seidenberg1989distributed} and suggests that the proposed learning never ceases \citep[see also][for studies demonstrating the effects of life-long learning]{ramscar2014myth, ramscar2017mismeasurement}. Our lexical knowledge continues to grow, and changes not only across a life-time, but also locally, from word use to word use.
Continuous recalibration is not restricted to language, but also takes place in, for instance,  vision, as demonstrated by the anti-priming effects studied by \citet{marsolek2008antipriming}.
Secondly, our results dovetail well with the general methodological law that a measurement instrument changes what it is designed to measure.  The lexical decision task likewise does not simply probe participants' lexical knowledge, it also changes this knowledge.  \citet{luce1995four} argued many years ago that psychological models need to move from being static to being dynamic.  We have shown that considerable headway can be made with incremental learning, as implemented in the DLM model.

Third, in addition to effects of trial-to-trial learning, in this study we also found that the
DLM-derived measures contributed substantial increases in model fit. Model fits improved more for the predictors derived from the dynamic models that incorporated trial-to-trial learning, compared to the static DLM models.  But even the predictors derived from the static model made it possible to substantially improve on models with the classical lexical predictors frequency, word length, and orthographic neighborhood size.
This finding adds to previous studies demonstrating the ability of the DLM to provide additional prediction accuracy for behavioural data \citep{chuang2021discriminative,Gahl2022thyme, schmitz2021durational, stein2021morpho}.  Of specific theoretical interest is that several of the DLM measures are grounded in a feedback loop \citep{chuang2020processing} from the semantics back to form. For modeling speech production, the DLM also includes a feedback loop from form to meaning.  The present simulation results therefore contribute evidence against strictly `feed-forward' models of lexical processing, and evidence in favor of models in which comprehension and production are to some extent interleaved.

Fourth, the predictors grounded in discriminative learning also provide more detailed insight into individual differences.
That individual differences exist between speakers is by itself unsurprising.  Language users have different exposure to and experience with language \citep{Gardner:Rothkopf:87a,hernandez2021german, Keuleers:2015, ramscar2016learning}.  Furthermore, cognitive differences may affect lexical processing \citep[e.g.][]{fischer2018individual, kuperman2011individual,loo2019effects, milin2017learning, perfetti2005word}.  As a consequence, the regression weights of lexical-distributional predictors can vary significantly between participants \citep[example in][]{baayen2014multivariate}. For the British Lexicon Project, we observed that faster subjects optimized their responses by focusing on meaning for words (leading to delays from well-supported forms), and on form for nonwords.  Slower subjects, by contrast,  optimized their responses by focusing on form across both words and nonwords. Furthermore, for nonwords, slower subjects suffered from interference from semantic neighbors.  This variegated pattern of response strategies for words and nonwords by variables of form and meaning is not detectable with the classical lexical variables word frequency, word length, and neighborhood density.  However, we expect  the Orthography-Semantics Consistency measure of \citet{marelli2015semantic} to provide further evidence for a differentiated role of semantics in the BLP lexical decision latencies, similar to the measure of semantic density that we used.

Fifth, the present study supports the possibility that nonwords are not totally devoid of meaning.  Earlier studies already presented experimental evidence for this possibility \citep{cassani2020semantics,chuang2020processing}. The present study adds to this an incremental perspective, in two ways. Firstly, especially slower responders took more time to reject nonwords when their predicted semantic vectors landed in a densely populated region of semantic space, and also when the predicted semantic vectors of their orthographic neighbors were spread out more widely in semantic space.  Secondly, we modeled the semantics associated with the lexical category of `nonword' as a continuously updated and ever changing location in semantic space, different across subjects, and within subjects updated from nonword trial to nonword trial in such a way that more recent nonword vectors were weighted more heavily. This implementation is in line with recency effects found in category learning \citep[e.g.][]{jones2006recency}.\footnote{For primacy effects in category induction, see \citet{duffy2008primacy}.}  This highly dynamic and continuously evolving meaning of the nonword category was successful in driving the error for nonwords in trial-to-trial incremental learning.

%%%%We modeled nonword semantics as a continuously updated and ever changing location in semantic space, different across subjects, and within subjects updated from nonword trial to nonword trial. %However, many other approaches are certainly possible, such as choosing the closest existing semantic vector as target, but an exploration of this issue was left to future work.

Since our main research interest in this study was the detection of continuous changes of lexical knowledge, we did not implement a decision process. This approach differs from many previous computational models of lexical decision which generally tried to implement a decision directly as part of the model. For example, architectures based on the interactive activation model \citep{rumelhart1982interactive} such as \citet{grainger1996orthographic}'s Multiple Read-out Model use the activations of individual word nodes to inform the decision process. \citet{norris2006bayesian}'s Bayesian reader implements a lexical decision mechanism based on the integration of a word's prior probability with incoming evidence. Similar to the Bayesian Reader, the DIANA model \citep{ten2022diana}, a model of auditory word recognition, implements different decision mechanisms depending on the task at hand. For lexical decision tasks, the activations between the highest supported word (modulated by prior probabilities, i.e. words' frequencies) and pseudoword candidates are compared, until they differ by some threshold $\theta$. Instead of incorporating a decision mechanism directly into our model, we adopted a two-step approach. The first step is the lexical processing of the incoming stimulus, which includes a comprehension mapping from form to meaning as well as a production mapping from meaning to form, underlining the integrated nature of the word recognition process \citep{chuang2020processing, liberman1985motor, pulvermuller2006motor}. Lexical processing is followed by the decision which is driven by general cognitive control processes. This approach was motivated in part by our research goal, showing that incremental discriminative learning can capture human trial-to-trial learning, but in part also by the work of \citet{Redgrave:Prescott:Gurney:1999} and \citet{Gurney:Prescott:Redgrave:2001}, who argue that decisions are made by distinct general cognitive control processes. Therefore, we made use of statistical models to establish the relative importance of various DLM-based lexical processing measures for lexical decision reaction times. However, improved insights into incremental lexical learning might be obtainable when the higher-order processes involved in lexicality decision making are modeled and allowed to feed back into the low-level processes of incremental lexical learning. The modeling of these processes is beyond the scope of the present study.\footnote{
Although reinforcement learning holds great promise for many areas of human decision making \citep{wilson2019ten}, it is unclear to us how appropriate this machine learning method is for the present data:  participants did not receive feedback on their choice behavior.  Furthermore, several observations on human response behavior in lexical decision tasks suggest that the Markov property, on which reinforcement learning builds, does not hold.  First, participants tend to be more careful after realizing they made an error \citep{laming1979autocorrelation}.  Second, participants are aware of  longer sequences consisting of only words or only nonwords, and may optimize their behavior accordingly.  Third, participants' behavior is subject to fluctuations in attention, which give rise to inter-trial dependencies that likely are inconsistent with rational decision making.
}

As is common in many computational modelling studies, we had to adopt a number of simplifications to avoid a combinatorial explosion of modelling decisions and to make our simulations feasible. For example, we chose a learning rate based on the data of one subject and used it across all subjects, even though individual differences in learning rate are to be expected \citep[e.g.][]{ezzizi2023error}. Furthermore, our model is based only on mappings between orthographic form and meaning and does not take into account any influences of mappings to phonology commonly assumed to be part of the reading process \citep[e.g.][]{amenta2017sound, newman2012does}. Future work should explore these aspects in more detail.

We conclude with a note on the Rescorla-Wagner and Widrow-Hoff learning rules.  The learning rule of Rescorla and Wagner has been used successfully in many areas of language-related research \citep[e.g.][]{ ellis2006language, ellis2006selective,nixon2021prediction,ramscar2013error}, including studies on trial-to-trial learning \citep{chuang2021discriminative, lentz2021temporal, tomaschek2022keys}. The learning rule of Widrow and Hoff has had much less impact in language and psychology, perhaps unsurprisingly, given the widespread use of discrete symbolic representations, especially for meanings and semantic features.   But with the advent of distributional semantics, and the widespread availability of high-quality word embeddings, the learning rule of Widrow and Hoff now comes into its own.  As demonstrated by the present study, it has exactly the right flexibility for trial-to-trial learning.  A challenge for further research is the incorporation of more powerful algorithms from deep learning, while retaining this flexibility of learning. 

\section*{Acknowledgments}
\noindent
Funded by the Deutsche Forschungsgemeinschaft (DFG, German Research Foundation) under Germany’s Excellence Strategy – EXC number 2064/1 – Project number 390727645, and by the European Research Council,  project WIDE-742545. The authors thank Fabian Tomaschek and Tino Sering for comments on earlier versions of this manuscript and participants of the Groningen Spring School on Cognitive Modeling for their helpful feedback.

\bibliography{joint_bib}

\end{document}

%% file: fig/ld_flowchart.tikz
\tikzset{every picture/.style={line width=0.75pt}} %set default line width to 0.75pt        

\begin{tikzpicture}[x=0.75pt,y=0.75pt,yscale=-1,xscale=1]
%uncomment if require: \path (0,484); %set diagram left start at 0, and has height of 484

%Curve Lines [id:da2656819545763036] 
\draw    (598,236.5) .. controls (593,280.5) and (485,420.5) .. (332,422.5) ;
\draw [shift={(332,422.5)}, rotate = 359.25] [color={rgb, 255:red, 0; green, 0; blue, 0 }  ][line width=0.75]    (10.93,-3.29) .. controls (6.95,-1.4) and (3.31,-0.3) .. (0,0) .. controls (3.31,0.3) and (6.95,1.4) .. (10.93,3.29)   ;
%Curve Lines [id:da3794805098261086] 
\draw    (198,235.5) .. controls (198,281.27) and (188.1,349.81) .. (248.09,423.39) ;
\draw [shift={(249,424.5)}, rotate = 230.5] [color={rgb, 255:red, 0; green, 0; blue, 0 }  ][line width=0.75]    (10.93,-3.29) .. controls (6.95,-1.4) and (3.31,-0.3) .. (0,0) .. controls (3.31,0.3) and (6.95,1.4) .. (10.93,3.29)   ;
%Straight Lines [id:da8303262947777852] 
\draw    (196,180.5) -- (196,100.5) ;
\draw [shift={(196,98.5)}, rotate = 90] [color={rgb, 255:red, 0; green, 0; blue, 0 }  ][line width=0.75]    (10.93,-3.29) .. controls (6.95,-1.4) and (3.31,-0.3) .. (0,0) .. controls (3.31,0.3) and (6.95,1.4) .. (10.93,3.29)   ;

% Text Node
\draw    (11,180) -- (79,180) -- (79,234) -- (11,234) -- cycle  ;
\draw (45,207) node   [align=left] {\begin{minipage}[lt]{43.52pt}\setlength\topsep{0pt}
\begin{center}
stimulus (string)
\end{center}

\end{minipage}};
% Text Node
\draw    (164,181) -- (232,181) -- (232,235) -- (164,235) -- cycle  ;
\draw (198,208) node   [align=left] {\begin{minipage}[lt]{43.52pt}\setlength\topsep{0pt}
\begin{center}
form vector $\displaystyle \mathbf{c}_t$
\end{center}

\end{minipage}};
% Text Node
\draw    (390,179.75) -- (525,179.75) -- (525,236.75) -- (390,236.75) -- cycle  ;
\draw (457.5,208.25) node   [align=left] {\begin{minipage}[lt]{89.08pt}\setlength\topsep{0pt}
\begin{center}
predicted semantic vector $\displaystyle \hat{\mathbf{s}}_t$
\end{center}

\end{minipage}};
% Text Node
\draw    (238,268.75) -- (373,268.75) -- (373,325.75) -- (238,325.75) -- cycle  ;
\draw (305.5,297.25) node   [align=left] {\begin{minipage}[lt]{89.08pt}\setlength\topsep{0pt}
\begin{center}
predicted form vector $\displaystyle \hat{\mathbf{c}}_t$
\end{center}

\end{minipage}};
% Text Node
\draw    (138,41.75) -- (253,41.75) -- (253,99.75) -- (138,99.75) -- cycle  ;
\draw (195.5,70.75) node   [align=left] {\begin{minipage}[lt]{75.48pt}\setlength\topsep{0pt}
\begin{center}
word/nonword outcome $\displaystyle \hat{\mathbf{d}}_t$
\end{center}

\end{minipage}};
% Text Node
\draw    (252,395) -- (330,395) -- (330,449) -- (252,449) -- cycle  ;
\draw (291,422) node   [align=left] {\begin{minipage}[lt]{50.32pt}\setlength\topsep{0pt}
\begin{center}
DLM measures
\end{center}

\end{minipage}};
% Text Node
\draw    (593,180) -- (661,180) -- (661,234) -- (593,234) -- cycle  ;
\draw (627,207) node   [align=left] {\begin{minipage}[lt]{43.52pt}\setlength\topsep{0pt}
\begin{center}
semantic vector $\displaystyle \mathbf{s}_t$
\end{center}

\end{minipage}};
% Text Node
\draw (118,207) node   [align=left] {\begin{minipage}[lt]{24.48pt}\setlength\topsep{0pt}
\begin{center}
turn \\into
\end{center}

\end{minipage}};
% Text Node
\draw (314,208) node   [align=left] {\begin{minipage}[lt]{96.56pt}\setlength\topsep{0pt}
\begin{center}
map to semantics using $\displaystyle \mathbf{F}_t$
\end{center}

\end{minipage}};
% Text Node
\draw (428,266) node   [align=left] {\begin{minipage}[lt]{96.56pt}\setlength\topsep{0pt}
\begin{center}
map back to \\form using $\displaystyle \mathbf{G}_t$
\end{center}

\end{minipage}};
% Text Node
\draw (172,325) node   [align=left] {\begin{minipage}[lt]{54.4pt}\setlength\topsep{0pt}
\begin{center}
use to \\calculate
\end{center}

\end{minipage}};
% Text Node
\draw (332,358) node   [align=left] {\begin{minipage}[lt]{54.4pt}\setlength\topsep{0pt}
\begin{center}
use to \\calculate
\end{center}

\end{minipage}};
% Text Node
\draw (557,347) node   [align=left] {\begin{minipage}[lt]{54.4pt}\setlength\topsep{0pt}
\begin{center}
use to \\calculate
\end{center}

\end{minipage}};
% Text Node
\draw (283.5,137) node   [align=left] {\begin{minipage}[lt]{116.28pt}\setlength\topsep{0pt}
map to word/nonword outcome using $\displaystyle \mathbf{D}_t$
\end{minipage}};
% Text Node
\draw (116.5,176.75) node  [color={rgb, 255:red, 155; green, 155; blue, 155 }  ,opacity=1 ] [align=left] {\begin{minipage}[lt]{17pt}\setlength\topsep{0pt}
(A)
\end{minipage}};
% Text Node
\draw (301.5,178.75) node  [color={rgb, 255:red, 155; green, 155; blue, 155 }  ,opacity=1 ] [align=left] {\begin{minipage}[lt]{17pt}\setlength\topsep{0pt}
(B)
\end{minipage}};
% Text Node
\draw (274.5,106.75) node  [color={rgb, 255:red, 155; green, 155; blue, 155 }  ,opacity=1 ] [align=left] {\begin{minipage}[lt]{17pt}\setlength\topsep{0pt}
(C)
\end{minipage}};
% Text Node
\draw (419.5,293.75) node  [color={rgb, 255:red, 155; green, 155; blue, 155 }  ,opacity=1 ] [align=left] {\begin{minipage}[lt]{17pt}\setlength\topsep{0pt}
(D)
\end{minipage}};
% Text Node
\draw (265.5,379.75) node  [color={rgb, 255:red, 155; green, 155; blue, 155 }  ,opacity=1 ] [align=left] {\begin{minipage}[lt]{17pt}\setlength\topsep{0pt}
(E)
\end{minipage}};
% Connection
\draw    (79,207.22) -- (162,207.76) ;
\draw [shift={(164,207.78)}, rotate = 180.37] [color={rgb, 255:red, 0; green, 0; blue, 0 }  ][line width=0.75]    (10.93,-3.29) .. controls (6.95,-1.4) and (3.31,-0.3) .. (0,0) .. controls (3.31,0.3) and (6.95,1.4) .. (10.93,3.29)   ;
% Connection
\draw    (232,208.03) -- (388,208.18) ;
\draw [shift={(390,208.18)}, rotate = 180.06] [color={rgb, 255:red, 0; green, 0; blue, 0 }  ][line width=0.75]    (10.93,-3.29) .. controls (6.95,-1.4) and (3.31,-0.3) .. (0,0) .. controls (3.31,0.3) and (6.95,1.4) .. (10.93,3.29)   ;
% Connection
\draw    (408.83,236.75) -- (355.9,267.74) ;
\draw [shift={(354.17,268.75)}, rotate = 329.65] [color={rgb, 255:red, 0; green, 0; blue, 0 }  ][line width=0.75]    (10.93,-3.29) .. controls (6.95,-1.4) and (3.31,-0.3) .. (0,0) .. controls (3.31,0.3) and (6.95,1.4) .. (10.93,3.29)   ;
% Connection
\draw    (302.19,325.75) -- (294.37,393.01) ;
\draw [shift={(294.14,395)}, rotate = 276.63] [color={rgb, 255:red, 0; green, 0; blue, 0 }  ][line width=0.75]    (10.93,-3.29) .. controls (6.95,-1.4) and (3.31,-0.3) .. (0,0) .. controls (3.31,0.3) and (6.95,1.4) .. (10.93,3.29)   ;

\end{tikzpicture}

%% file: fig/shortest_path_diagram.tikz
\tikzset{every picture/.style={line width=0.75pt}} %set default line width to 0.75pt        

\begin{tikzpicture}[x=0.75pt,y=0.75pt,yscale=-1,xscale=1]

%Shape: Circle [id:dp697619947647234] 
\draw  [fill={rgb, 255:red, 0; green, 0; blue, 0 }  ,fill opacity=1 ] (127.56,178.92) .. controls (125.8,180.68) and (122.95,180.68) .. (121.19,178.92) .. controls (119.44,177.17) and (119.44,174.32) .. (121.19,172.56) .. controls (122.95,170.8) and (125.8,170.8) .. (127.56,172.56) .. controls (129.31,174.32) and (129.31,177.17) .. (127.56,178.92) -- cycle ;
%Shape: Circle [id:dp7511067684884152] 
\draw  [fill={rgb, 255:red, 0; green, 0; blue, 0 }  ,fill opacity=1 ] (163.94,211.98) .. controls (161.46,212.22) and (159.26,210.41) .. (159.02,207.94) .. controls (158.78,205.46) and (160.59,203.26) .. (163.06,203.02) .. controls (165.53,202.78) and (167.74,204.59) .. (167.98,207.06) .. controls (168.22,209.53) and (166.41,211.74) .. (163.94,211.98) -- cycle ;
%Shape: Circle [id:dp45507203888849135] 
\draw  [fill={rgb, 255:red, 126; green, 211; blue, 33 }  ,fill opacity=1 ] (274,135.41) .. controls (274.05,132.93) and (276.1,130.95) .. (278.59,131) .. controls (281.07,131.05) and (283.05,133.1) .. (283,135.59) .. controls (282.95,138.07) and (280.9,140.05) .. (278.41,140) .. controls (275.93,139.95) and (273.95,137.9) .. (274,135.41) -- cycle ;
%Shape: Circle [id:dp21909297285322638] 
\draw  [fill={rgb, 255:red, 0; green, 0; blue, 0 }  ,fill opacity=1 ] (237.37,37.54) .. controls (239.56,36.36) and (242.29,37.18) .. (243.46,39.37) .. controls (244.64,41.56) and (243.82,44.29) .. (241.63,45.46) .. controls (239.44,46.64) and (236.71,45.82) .. (235.54,43.63) .. controls (234.36,41.44) and (235.18,38.71) .. (237.37,37.54) -- cycle ;
%Straight Lines [id:da47766843363033884] 
\draw    (241.63,45.46) -- (278.59,131) ;
%Straight Lines [id:da8250805110270639] 
\draw    (167.98,207.06) -- (276.15,139.34) ;
%Straight Lines [id:da6737419507858802] 
\draw    (127.56,178.92) -- (159.32,205.84) ;
%Straight Lines [id:da05603430259584419] 
\draw    (125.4,171.36) -- (235.54,43.63) ;
%Straight Lines [id:da15491092793656358] 
\draw    (128.76,176.76) -- (274,135.41) ;
%Straight Lines [id:da6281561567769488] 
\draw    (165.16,203.32) -- (237.26,45.4) ;

% Text Node
\draw (133,194) node [anchor=north west][inner sep=0.75pt]  [font=\small] [align=left] {1};
% Text Node
\draw (262,79) node [anchor=north west][inner sep=0.75pt]  [font=\small] [align=left] {2};
% Text Node
\draw (225,171) node [anchor=north west][inner sep=0.75pt]  [font=\small] [align=left] {2.5};
% Text Node
\draw (153,102) node [anchor=north west][inner sep=0.75pt]  [font=\small] [align=left] {3.5};
% Text Node
\draw (246,29) node [anchor=north west][inner sep=0.75pt]   [align=left] {sack};
% Text Node
\draw (88,166) node [anchor=north west][inner sep=0.75pt]   [align=left] {tack};
% Text Node
\draw (153.02,210.94) node [anchor=north west][inner sep=0.75pt]   [align=left] {lack};
% Text Node
\draw (285,128) node [anchor=north west][inner sep=0.75pt]   [align=left] {back};
% Text Node
\draw (213,95) node [anchor=north west][inner sep=0.75pt]  [font=\small] [align=left] {4};
% Text Node
\draw (219,135) node [anchor=north west][inner sep=0.75pt]  [font=\small] [align=left] {3};

\end{tikzpicture}